%% file: main.tex
\documentclass[10pt,twocolumn,letterpaper]{article}

\usepackage{cvpr}
\usepackage{times}
\usepackage{epsfig}
\usepackage{graphicx}
\usepackage{amsmath}
\usepackage{amssymb}

\usepackage{subfigure} 
\usepackage{amsfonts} 
\usepackage{placeins}
\usepackage{algorithm}
\usepackage{algorithmic}
\usepackage{array}
\usepackage{overpic}
\usepackage{wrapfig}
\usepackage{bm}
\usepackage{comment}

\usepackage{pgfplots}\pgfplotsset{compat=newest}
\newlength\figureheight 
\newlength\figurewidth 
\usetikzlibrary{plotmarks}
\usepackage{pgf}
\usepackage{tikz}
\usepackage{tkz-euclide}
\usepackage{tkz-graph}
\usetikzlibrary{calc}
\usetkzobj{all}
\usepackage{soul}
\usepackage{appendix}

\providecommand{\keywords}[1]
{
  \small	
  \textbf{Keywords: } #1
}

\newcommand{\norm}[1]{\left\lVert#1\right\rVert}

\usepackage{hyperref}
\hypersetup{pagebackref=true,breaklinks=true,colorlinks,bookmarks=false}

\cvprfinalcopy 

\def\cvprPaperID{5649} 

\ifcvprfinal\fi
\begin{document}

\title{The Whole Is Greater Than the Sum of Its Nonrigid Parts\\
\includegraphics[width=0.9\textwidth]{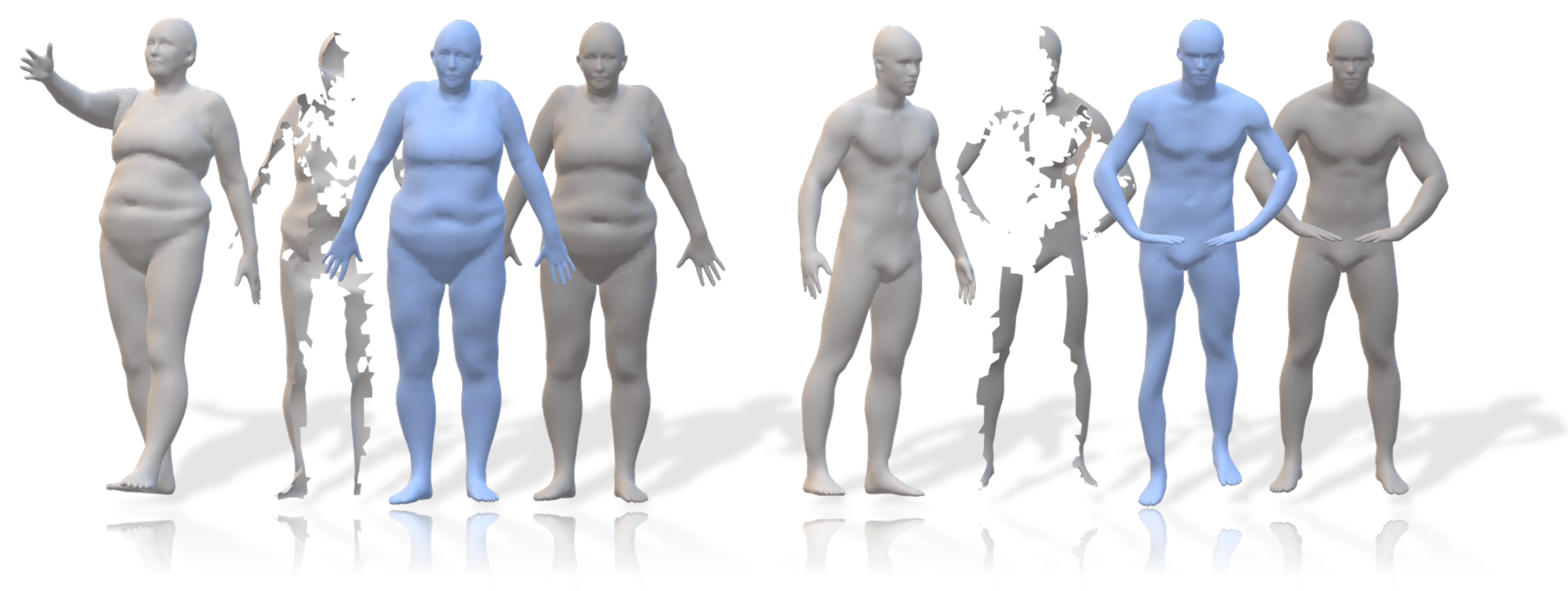}\\
{\normalfont \normalsize Left to right: input reference shape, input part, output  completion, and the ground truth full model.}
}

\author{Oshri Halimi\\
Technion, Israel\\
{\tt\small oshri.halimi@gmail.com}
\and
Ido Imanuel\\
Technion, Israel\\
{\tt\small ido.imanuel@gmail.com}
\and
Or Litany\\
Stanford University\\
{\tt\small or.litany@gmail.com}
\and
Giovanni Trappolini\\
Sapienza University of Rome\\
{\tt\small giovanni.trappolini@uniroma1.it}
\and
Emanuele Rodol\`{a}	\\
Sapienza University of Rome\\
{\tt\small rodola@di.uniroma1.it}
\and
Leonidas Guibas\\
Stanford University\\
{\tt\small guibas@cs.stanford.edu}
\and
Ron Kimmel\\
Technion, Israel\\
{\tt\small ron@cs.technion.ac.il}
}

\maketitle
\begin{abstract}
 According to Aristotle, a philosopher in Ancient Greece, {\it ``the whole is greater than the sum of its parts''}. 
 This observation was adopted to explain human perception by the Gestalt psychology school of thought in the twentieth century. Here, we claim that observing part of an object which was previously acquired as a whole, one could deal with both partial matching and 
 shape completion in a holistic manner. 
 More specifically, given the geometry of a full, articulated object in a given pose, as well as a partial scan of the same object in a different pose, we address the problem of matching the part to the whole while simultaneously reconstructing the new pose from its partial observation. 
 Our approach is data-driven, and takes the form of a Siamese autoencoder without the requirement of a consistent vertex labeling at inference time; as such, it can be used on unorganized point clouds as well as on triangle meshes. 
 We demonstrate the practical effectiveness of our model in the applications of single-view deformable shape completion and dense shape correspondence, both on synthetic and real-world geometric data, where we outperform prior work on these tasks by a large margin.
 
\end{abstract}


\section{Introduction}
\input{sections/intro.tex}
\section{Related work}
\input{sections/relatedwork.tex}

\section{Method}
\input{sections/method.tex}
\section{Experiments}
\input{sections/experiments.tex}
\section{Concluding remarks}
\input{sections/conclusions.tex}

\section*{Appendix}
\appendix
\input{sections/Supplementary.tex}

{\small
\bibliographystyle{ieee_fullname}
\bibliography{main}
}

\end{document}

%% file: sections/intro.tex
 Aristotle, a philosopher in Ancient Greece declared that
 {\it ``the whole is greater than the sum of its parts''}. 
 This fundamental observation was narrowed down to human perception of planar shapes by the Gestalt psychology school of thought in the twentieth century.
 A guiding idea of Gestalt theory is the principle of {\em reification}, arguing that human perception contains more spatial information than can be extracted from the sensory stimulus, and thus giving rise to the view that the mind generates the additional information based on verbatim acquired patterns.
Here we adopt this line of thought in the context of non-rigid shape completion. Specifically, we argue that given access to a complete shape in one pose, one can accurately complete partial views of that shape at any other pose. 

Shape completion is a problem of great practical importance; Advanced applications in robotics as well as augmented and virtual reality often rely upon the ability to render a scene from novel views, manipulate its content, and add physical constraints. 
This requires the completion of geometric structures from partial data.
To that end, we would like to have a way to complete a partial shape into its full counterpart.

Shape completion has been studied in previous papers in the setting where only access to a partial input is given. 
This renders the problem ill-posed by definition as there can be many plausible completions to a given input. 
As an example, consider observing the rear part of a human shape and trying to infer the face; Generally, in this case, there is no unique  answer. 
This plain observation is reflected in the literature; While completion methods operating only on the partial shape \cite{litany2017deformable, marin2018farm} show high capabilities of pose reconstruction, these methods suffer greatly from poor reconstruction of the fine details distinguishing one human subject from another.

In this paper, we introduce a {\em deterministic} shape completion framework. 
For a partial observation, the method returns a reliable reconstruction of a full, realistic object.

At a first glance, at least in a rigid setting, this task may seem impossible: How can we guarantee that such a reconstruction reliably describes the part hidden from the viewing direction? 
However, one soon realizes that the problem can be formulated differently in the {\em non-rigid} case.

Since two non-rigidly related shapes could share the same intrinsic geometry \cite{EladKimmel2003,BBK2007_IEEIP,BBK2005_IJCV, halimia2019computable, halimi2018self}, each shape holds information about the other. 
We therefore pose the alternative question: Given a full object $Q$ and a partial view $P$ {\em in a different pose}, can we reconstruct a {\em full} version of $P$ by borrowing geometric information from $Q$? In this paper, we address precisely this question. 
While our method requires access to the full model, it is a reasonable assumption for certain applications that one would first undergo a complete scan in one pose to enable further accurate reconstructions in arbitrary poses. 
However, though theoretically interesting and of great practical importance this problem, until now, was not addressed in the literature.

We solve this problem by deforming the full shape $Q$ to align with the partial shape $P$, hereby producing the completion as well as the partial shape correspondence. We propose a learning scheme that predicts the deformation based on two different sources of geometric information; The partial shape $P$ determines the pose,  and the full shape $Q$, encapsulates the detailed geometric information identifying the subject.
We note that the most common setting of partiality is when a subject is acquired from a single view. Therefore, we focus on this case where the pose, in principle, can be extracted from the partial observation $P$. We also note that partial surface overlapping is generally a stricter constraint than can be imposed by skeleton measurements alone, therefore we are interested in this exact setting as described above.

Our goal can thus be seen as a combination of two complementary tasks: (1) partial shape matching, and (2) non-rigid shape completion. While the two tasks are often addressed separately, we claim that considering their coupling provides powerful means to deal with both.

\paragraph*{Contribution.}
In this paper we propose a novel formulation unifying deformable shape completion and partial shape matching.
Specifically, given a full and a partial shape related by a non-rigid transformation, our objective is to deform the full shape to best fit the partial data. 
To compute this deformation, we train an encoder-decoder network. Once a completion for the part is predicted, the part-to-full correspondence can be trivially recovered using nearest-neighbor search.
Our main contributions can be summarized as follows:

\begin{enumerate}
\item 
 We introduce a deep Siamese architecture to tackle non-rigid alignment between a shape and its partial scan;
\item 
To the best of our knowledge, the proposed method is the first that addresses shape completion and partial correspondence in one framework under {\em extreme} partiality;
\item 
The proposed method is {\em efficient}, taking less than a second to provide both outputs without requiring any time-consuming post-processing steps.
\end{enumerate}

%% file: sections/relatedwork.tex
Our problem setting is closely related to multiple research directions in the shape analysis and geometric deep learning communities.

\vspace{1ex}\noindent\textbf{Shape completion.}

Recovering a complete shape from partial or noisy measurements is a longstanding research problem that comes in many flavors. 
In an early attempt to use one pose in order to geometrically reconstruct another, Devir~\etal~\cite{Devir_NORDIA2009} considered mapping a model shape in a given pose onto a noisy version of the shape in a different pose. 
Elad and Kimmel were the first to treat shapes as metric spaces \cite{EladKimmel_CVPR2001,EladKimmel2003}. They matched shapes by comparing second order moments of embedding the intrinsic metric into a Euclidean one via classical scaling. 
In the context of deformable shapes, early efforts focused on completion based on geometric priors~\cite{kazhdan2013screened} or reoccurring patterns ~\cite{BBK_ECCV2006,korman2015peeking,sarkar2017learning,litany2016cloud}. 
These methods are not suited for severe partiality. 
For such cases model-based techniques are quite popular, for example, 
 category-specific parametric morphable models that can be fitted to the partial data~\cite{blanz1999morphable,gerig2017morphable,loper2015smpl,allen2006learning,shtern2018fast}. 
Model-based shape completion was demonstrated for key-points input~\cite{anguelov2005scape}, and was recently proven to be quite useful for recovering 3D body shapes from 2D images~\cite{varol2017learning,varol2018bodynet,guler2019holopose,zanfir2018monocular}. 
Parametric morphable models \cite{blanz1999morphable}, coupled with axiomatic image formation models were used to train a network to reconstruct face geometry from images \cite{Richardson_3DV2016,Richardson_CVPR2017,Sela_ICCV2017}.
Still, much less attention has been given to the task of fitting a model to a partial 3D point cloud. 
Recently, Jiang~\etal~\cite{jiang2019skeleton} tackled this problem using a skeleton-aware architecture. 
However, their approach works well when full coverage of the underlying shape is given.

\vspace{1ex}\noindent\textbf{Deep learning of surfaces.} 

Following the success of convolutional neural networks in images, in recent years, the geometry processing community has been rapidly adopting and designing computational modules suited for such data. 
The main challenge is that unlike images, geometric structures like surfaces come in many types of representations, and each requires a unique handling. 
Early efforts focused on a simple extension from a single image to multi-view representations~\cite{su2015multi,wei2016dense}. 
Another natural extension are 3D CNN on volumetric grids~\cite{wu20153d}. 
A host of techniques for mesh processing were developed as part of a research branch termed \textit{geometric deep learning}~\cite{bronstein2017geometric}. 
These include graph-based methods~\cite{dynFilt, wang2019dynamic, hanocka2019meshcnn}, intrinsic patch extraction~\cite{masci2015geodesic,boscaini2016learning,monet}, and spectral techniques~\cite{litany2017deep, halimi2019unsupervised}. 
Point cloud networks~\cite{qi2016pointnet,qi2017pointnet++} have recently gained much attention. 
Offering a light-weight computation restricted to sparse points with a sound geometric explanation \cite{Joseph-Rivlin_CVPR2019},
these networks have shown to provide a good compromise between complexity and accuracy, and are dominating the field of 3D object detection ~\cite{qi2019deep, xu2018pointfusion}, semantic segmentation~\cite{graham20183d,BenShabat_2017}, and even temporal point cloud processing~\cite{choy20194d,liu2019meteornet}.
For generative methods, recent implicit and parametric methods have demonstrated promising results~\cite{groueix2018atlasnet,park2019deepsdf}.

Following the success of encoding non-rigid shape deformations using a point cloud network~\cite{groueix20183d}, here, we also choose to use a point cloud representation. 
Importantly, while the approach presented in~\cite{groueix20183d} predicts alignment of two shapes, it is not designed to handle severe partiality, and assumes a fixed template for the source shape. 
Instead, we show how to align arbitrary input shapes and focus on such a partiality.

\vspace{1ex}\noindent\textbf{Partial shape matching.}
Dense non-rigid shape correspondence \cite{kim11,chen15,litany2017deep,halimi2019unsupervised, rodola2014dense,bronstein2006generalized,cosmo2019isospectralization} is a key challenge in 3D computer vision and graphics, and has been widely explored in the last few years. 
A particularly challenging setting arises whenever one of the two shapes has missing geometry.  
Bronstein \etal \cite{BBK_AVBPA2003,BBK2005_IJCV,BBK_ECCV2006,BBBK_AMDO2006,BBK_AMDO2006,BBK2007_IEEIP}  dealt with partial matching of articulated objects in various scenarios, including pruning of the intrinsic structure while accounting for cuts. 
This setting has been tackled with moderate success in a few recent papers \cite{rodola2017partial,litany2017fully,rampini2019correspondence}, however, it largely remains an open problem whenever the partial shape exhibits severe artifacts or large, irregular missing parts. 
In this paper we tackle precisely this setting, demonstrating unprecedented performance on a variety of real-world and synthetic datasets.

%% file: sections/method.tex
\begin{figure*}[ht]
    \centering
     \includegraphics[width=0.7\linewidth]{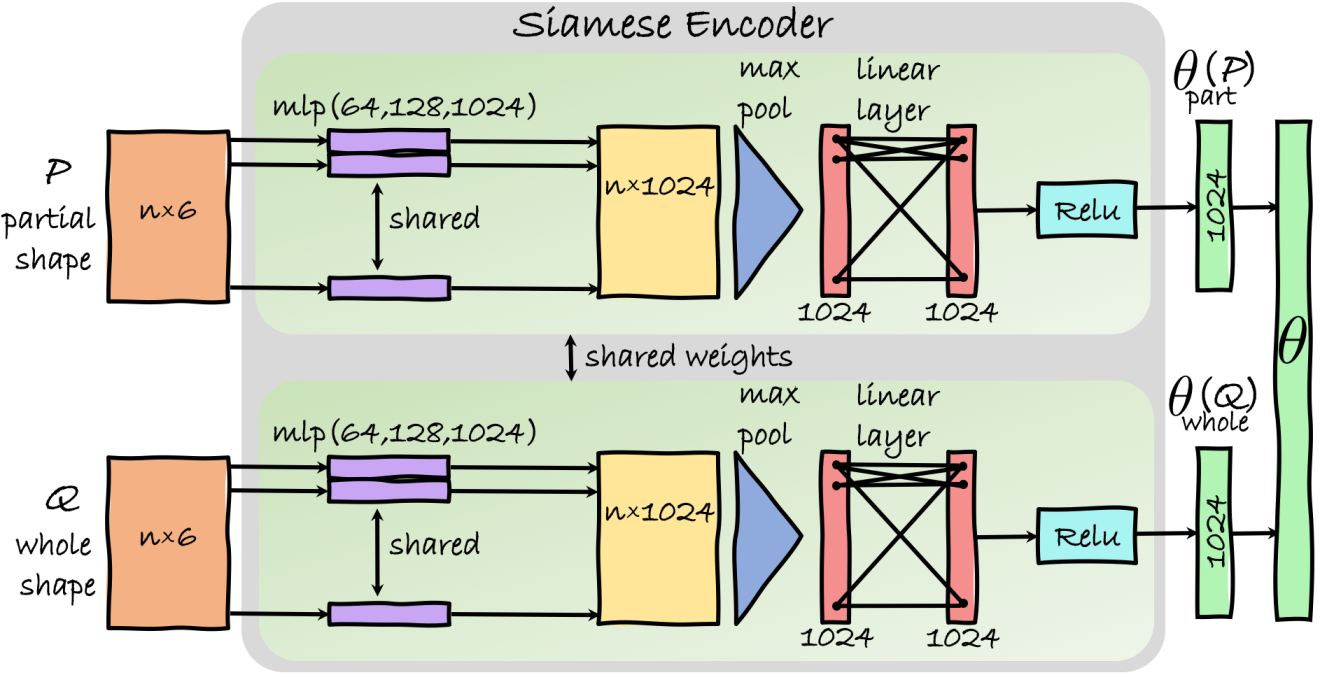}
     \includegraphics[width=0.7\linewidth]{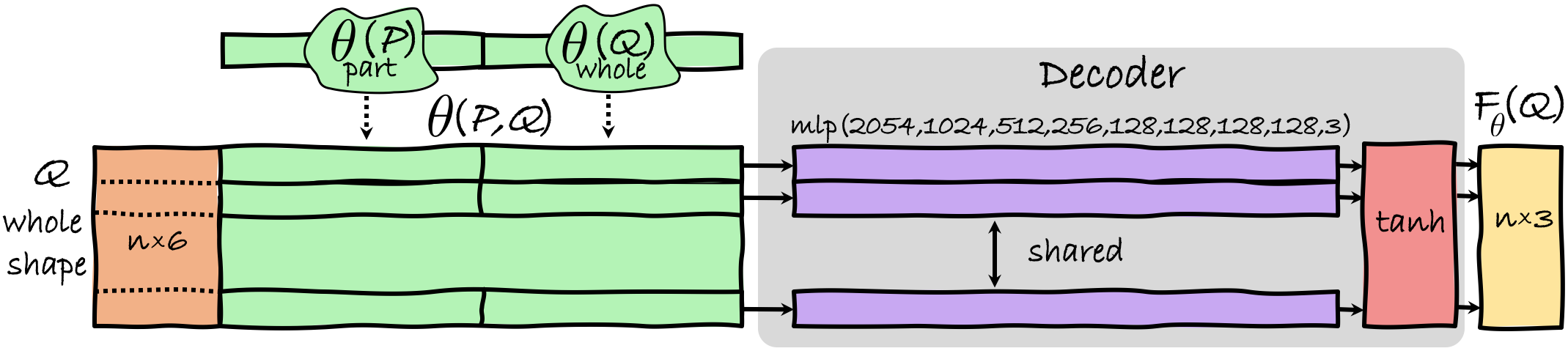}
    \caption{\textbf{Network Architecture}. Siamese encoder architecture at the top, and the decoder (generator) architecture at the bottom. A shape is provided to the encoder as a list of 6D points, representing the spatial and unit normal coordinates. The latent codes of the input shapes $\theta_{part}(P)$ and $\theta_{whole}(Q)$ are concatenated to form a latent code $\theta$ representing the input pair. Based on this latent code, the decoder deforms the full shape by operating on each of its points with the same function. The result is the deformed full shape $F_{\theta}$(Q).  
    }
    \label{fig:network_architecture}
\end{figure*}

\subsection{Overview}
We represent shapes as point clouds $S = \{s_i\}_{i=1}^{n_s}$ embedded in $\mathbb{R}^3$. 
Depending on the setting, each point may carry additional semantic or geometric information encoded as feature vectors in $R^d$. For simplicity we will keep d=3 in our formulation.

Given a full shape $Q  = \{q_i\}_{i=1}^{n_q}$ and its partial view in a different pose $P  = \{p_i\}_{i=1}^{n_p}$, our goal is to find a nonlinear function $F: \mathbb{R}^3 \rightarrow\mathbb{R}^3$ aligning $Q$ to $P$\footnote{In our setting, we assume that the pose can be inferred from the partial shape (\eg, an entirely missing limb would make the prediction ambiguous), hence the deformation function $F$ is well defined.}. If $R = \{r_i\}_{i=1}^{n_r}$ is the (unknown) full shape such that $P\subset R$, ideally we would like to ensure that $F(Q)=R$, where equality should be understood as same underlying surface. Thus, the deformed shape $F(Q)$ acts as a proxy to solve for the correspondence between the part $P$ and the whole $Q$. By calculating for every vertex in $P$ its nearest neighbor in $R \approx F(Q)$, we trivially obtain the mapping from $P$ to $Q$. 

The deformation function $F$ depends on the input pair of shapes $(P,Q)$. We model this dependency by considering a parametric function $F_{\theta}: \mathbb{R}^3 \rightarrow\mathbb{R}^3$, where $\theta$ is a latent encoding of the input pair $(P,Q)$. We implement this idea via an encoder-decoder neural network, and learn the space of parametric deformations from example pairs of partial and complete shapes, together with full uncropped version the partial shape, serving as the ground truth completion. 

Our network is composed of an encoder $E$ and a generator $F_\theta$. The encoder takes as input the pair $(P, Q)$ and embeds it into a latent code $\theta$. To map points from $Q$ to their new location, we feed them to the generator along with the latent code. 
Our network architecture shares a common factor with 3D-CODED architecture \cite{groueix20183d}, namely the deformation of one shape based on the latent code of the another. However \cite{groueix20183d} uses a fixed template and is therefore only suited for no- or mild-partiality as the template cannot make up for lost shape details in the part. Our pipeline on the other hand, is designed to merge two sources of information into the reconstructed model, resulting in an accurate reconstruction under extreme partiality. In Appendix \ref{fixed_template} we perform an analysis where we train our network in a fixed-template setting, similar to 3D-CODED and demonstrate the advantage of our paradigm. 

In what follows we first describe each module, and then give details on the training procedure and the loss function. We refer to Figure~\ref{fig:network_architecture} for a schematic illustration of our learning model.

\subsection{Encoder}
We encode $P$ and $Q$ using a Siamese pair of single-shape encoders, each producing a global shape descriptor (respectively $\theta_{part}$ and $\theta_{whole}$). The two codes are then concatenated so as to encode the information of the specific pair of shapes, $\theta = [\theta_{part}, \theta_{whole}]$.

Considering the specific architecture of the single-shape encoder, we think about the encoder network as a channel transforming geometric information to a vector representation. We would like to utilize architectures which have been empirically proven to encode the 3D surface with the least loss of information, thus enabling the decoder to convert the resulting latent code $\theta$ to an accurate spatial deformation $F_{\theta}$. Encouraged by recent methods \cite{groueix2018atlasnet,groueix20183d} that showed detailed reconstruction using PointNet \cite{qi2016pointnet}, we also adopt it as our backbone encoder. Specifically, our encoder passes all 3D points of the input shape through the same multi-layer perceptron with hidden dimensions $(64, 128,1024)$. A max-pool operation over the input points, then leads to a single $1024$-dimensional vector. Finally, we apply a linear layer of size $1024$ and a ReLu activation function. Hence, each shape in the input pair is represented by a latent code $\theta_{whole}, \theta_{part}$ of size $1024$ respectively. We concatenate these to a joint representation $\theta$ of size $2048$. 

In practice we found it helpful to include normal vectors as additional input features, making each input point 6D. The normal vector field is especially helpful for disambiguating contact points of the surface and by which to prevent contradicting requirements of the estimated deformation function. The normals were computed using the connectivity for mesh inputs, and approximated from neighboring points for point clouds, as described in the experimental section. 

\subsection{Generator}
Given the code $\theta$, representing the partial and full shapes, the generator has to predict the deformation function $F_{\theta}$ to be applied to the full shape $Q$. 
We realize $F$ as a Multi-Layer Perceptron (MLP) that maps an input point $q_i$ on the full shape $Q$, to its corresponding output point $r_i$ on the ground truth completed shape. The MLP operates pointwise on the tuple $(q_i, \theta)$, with $\theta$ kept fixed. The result is the destination location $F_{\theta}(q_i) \in \mathbb{R}^3$, for each input point of the full shape $Q$. 
This generator architecture allows, in principle, to calculate the output reconstruction in a flexible resolution, by providing the generator a full shape with some desired output resolution. 
In detail, the generator consists of $9$ layers of hidden dimensions $(2054, 1024, 512, 256, 128, 128, 128, 128, 3)$, followed by a hyperbolic tangent activation function. The output of the decoder is the 3D coordinates. In addition, we can compute a normal field based on the vertex coordinates, making the overall output of the decoder a 6D point. The normal is calculated using the known connectivity or alternatively, if the training dataset consists of point clouds, the output coordinates could be further provided to a differentiable normal estimator~\cite{ben2019nesti, qi2016pointnet}.

\subsection{Training Procedure}

We train our model using samples from datasets of human shapes. These contain 3D models of different subjects in various poses.  The datasets are described in detail in Section \ref{Datasets}.  
Each training sample is a triplet $(P, Q, R)$ of a partial shape $P$, a full shape in a different pose $Q$ and a ground truth completion $R$. The shapes $Q$ and $R$ are sampled from the same subject in two different poses. To get $P$ we render a depth map from $R$ at a viewpoint of zero elevation and a random azimuth angle in the range $0^{\circ}$ and $360^{\circ}$. These projections approximate the typical partiality pattern of depth sensors. Note that despite the large missing region, these projections largely retain the pose, making the reconstruction task well-defined.
The training examples $(P_n, Q_n, R_n)_{n=1}^N$ were provided in batches to the Siamese Network, where $N$ is the size of the train set. 
Each input pair is fed to the encoder to receive the latent code $\theta(P_n, Q_n)$ and the reconstruction $F_{\theta(P_n,Q_n)}(Q_n)$ is determined by the generator. 
This reconstruction is subsequently compared against the ground-truth reconstruction $R_n$ using the loss defined in the next subsection \ref{Loss}.

\subsection{Loss function}
\label{Loss}

The loss definition should reflect the visual plausibility of the reconstructed shape. Measuring such a quality analytically is a challenging problem worth studying in itself. Yet, in this paper we adopt a naive measurement of the Euclidean proximity between the ground-truth and the reconstruction.
Formally, we define the loss as,
\begin{equation}\label{eq:loss}
    \mathcal{L}(P,Q,R) = \sum_{i=0}^{n_q}\norm{F_{\theta(P,Q)}(q_i) - r_i}^2,
\end{equation}

where $r_i = \pi^*(q_i) \in R$ is the matched point of $q_i \in Q$, given by the ground-truth mapping $\pi^* : Q \rightarrow R$.

In practice we found that predicting normal vectors in addition to point coordinates helps to preserve fine details. We supervise for it using equation \ref{eq:loss} by defining $r_i$ as the concatenation of the coordinates and unit normal vector at each point: $ (\vec x_{ri}, \alpha\vec n_{ri}) \in \mathbb{R}^6$.

\subsection{Implementation considerations}
Our implementation is available at \url{https://github.com/OshriHalimi/shape_completion}.
The network was trained with each batch containing $10$ triplet examples $(P, Q, R)$, using the PyTorch \cite{paszke2017automatic} ADAM optimizer with a learning rate of $0.001$ and a momentum of $0.9$. 
We used a scale factor of $\alpha = 0.1$ for the normal vector. 
The network was trained for $50$ epochs, each containing $10,000$ random triplet examples. 
The input shapes were translated such that their center of mass lies at the origin. To align the reconstructed shape with the partial input, we run a partial Iterative Closest Point algorithm \cite{besl1992method}. Recovering the partial correspondence with respect to the aligned reconstruction results in improved accuracy. 

%% file: sections/experiments.tex
The proposed method tackles two important tasks in nonrigid shape analysis: shape completion and partial shape matching. We emphasize the graceful handling of severe partiality resulting from range scans.  In contrast, prior efforts either addressed one of these tasks or attempted to address both at mild partiality conditions. 
In this section we first describe the different datasets used. We then test our method on both  tasks and compare with prior art. Finally, we show performance on real scanned data. 

\subsection{Datasets}
\label{Datasets}

We utilize two datasets of human shapes for training and evaluation, FAUST~\cite{faust} and AMASS~\cite{AMASS:2019}. In addition we use raw scans from Dynamic FAUST \cite{dfaust:CVPR:2017} for testing purposes only. 
FAUST was generated by fitting SMPL parametric body model~\cite{loper2015smpl} to raw scans. It is a relatively small set of $10$ subjects posing at $10$ poses each. Following training and evaluation protocols from previous works (e.g. \cite{litany2017deep}), we test our method on partial shape matching and shape completion tasks using $10$ projected views of a held-out test set from FAUST. AMASS, on the other hand, is currently the largest and most diverse dataset of human shapes designed specifically for deep learning applications. It was generated by unifying $15$ archived datasets of marker-based optical motion capture (mocap) data. Each mocap sequence was converted to a sequence of rigged meshes using SMPL+H model~\cite{MANO:SIGGRAPHASIA:2017}. We then turn to AMASS which provides a richer resource for evaluating generalization. We generated a large set of single-view projections by sampling every 100th frame of all provided sequences. We then rendered each shape from $10$ equally spaced azimuth angles (keeping elevation at zero) using pyRender \cite{huang2019framenet}. Keeping the data splits prescribed by~\cite{AMASS:2019}, our dataset comprises a total of $110K$, $10K$, and $1K$ full shapes for train, validation and test, respectively; and $10$ times many partial shapes. Note that at train time we randomly mix and match full shapes and their parts which drastically increases the effective set size. 

\subsection{Methods in comparison}
\label{baselines}

The problem of deformable shape completion was recently studied by Litany \etal~\cite{litany2017deformable}. In their work, completion is achieved via optimization in a learned shape space. Different from us, their task is completion from a partial view without explicit access to a full model. This is an important distinction as it means missing parts can only be hallucinated. In contrast, we assume the shape details are provided but are not in the correct pose. Moreover, their solution requires a preliminary step of solving partial matching to a template model, which itself is a hard problem. Here, we solve for it together with the alignment. The optimization at inference time also makes their solution quite slow. Instead we output a result in a single feed forward fashion. 3D-CODED~\cite{groueix20183d} performs template alignment to an input shape in two stages: fast inference and slow refinement. It is designed for inputs which are either full or has mild partiality. Here we evaluate the performance of their network predictions under significant partiality. In the refinement step we use directional Champfer distance, as suggested by the authors in the partial case. 
FARM~\cite{marin2018farm} is another alignment-based solution that has shown impressive results on shape completion and dense correspondences. It builds on the SMPL~\cite{loper2015smpl} human body model due to its compact parameterization, yet, we found it to be very slow to converge (up to $30$ min for a single shape) and prone to getting trapped in local minima. 
3D-EPN~\cite{dai2016shape} is a rigid shape completion method. Based on a 3D-CNN, it accepts a voxelized signed distance field as input, and outputs that of a completed shape. Results are then converted to a mesh via computation of an isosurface.
Comparison with classic Poisson reconstruction~\cite{kazhdan2013screened} is also provided. It serves as a na\"ive baseline as it has access only to the partial input.

\subsection{Evaluation metrics}
\label{eval_metric}

Lacking a single good measure of completion quality, we provide $5$ different ones (see tables \ref{tab_proj_completion},\ref{tab:tab_amass_completion}). Each measurement highlights a different aspect of the predicted completion. MSE refers to the mean square error of the Euclidean distance between each point on the reconstructed shape and its ground truth mapping. We report MSE for predictions with well defined correspondence to the true reconstruction. We also report the MSE of two directional Chamfer distances: ground-truth to prediction, and vice versa. The former measures coverage of the target shape by the prediction and the later penalizes prediction outliers. We report the sum of both as full the Chamfer distance. Finally, we report volume deformation as the absolute volume difference divided by the ground truth volume.

\subsection{Single view completion}\label{subsec:shape_completion}
We evaluate our method on the task of deformable shape completion on FAUST and AMASS. 

\begin{figure}[htbp]
    \centering
    \includegraphics[width=\linewidth]{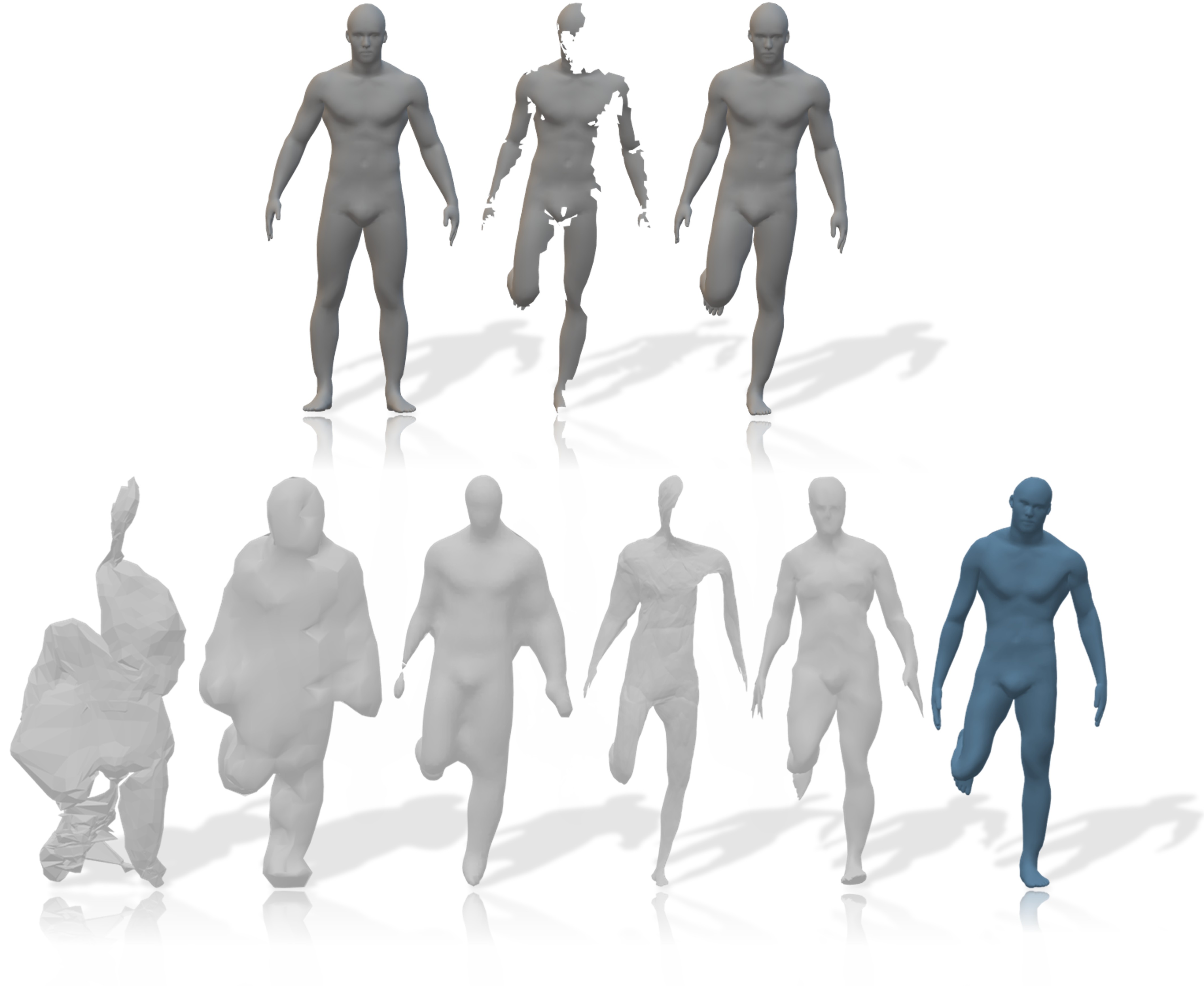}
    \caption{\textbf{FAUST Shape Completion}. At the top from left to right: full shape $Q$, partial shape $P$, ground truth completion $R$. At the bottom from left to right: reconstructions from FARM \cite{marin2018farm}, 3D-EPN \cite{dai2016shape}, Poisson \cite{kazhdan2013screened}, 3D-CODED \cite{groueix20183d}, Litany \textit{et al} \cite{litany2017deformable} and ours.}
    \label{fig:partial_matching_qualitative}
\end{figure}

\paragraph{FAUST projections}
We follow the evaluation protocol proposed in \cite{litany2017deformable} and summarize the completion results of our method and prior art in Table \ref{tab_proj_completion}. As can be seen, our network generates a much more accurate completion. Contrary to optimization-based methods \cite{litany2017deformable, groueix20183d, marin2018farm} which are very slow at inference, our feed-forward network performs inference in less than a second. To better appreciate the quality of our reconstructions, in Figure \ref{fig:partial_matching_qualitative} we visualize completions predicted by various methods. Additional results are provided in Appendix \ref{additional_vis}. Note how our method accurately preserves fine details that were lost in previous methods. 

\paragraph{AMASS projections}
Using our test set of partial shapes from AMASS (generated as described in \ref{Datasets}), we compare our method with two recent methods based on shape alignment: 3D-CODED~\cite{groueix20183d}, and FARM~\cite{marin2018farm}. As described in \ref{baselines}, 3D-CODED is a learning-based method that uses a fixed template and is not designed to handle severe partiality. FARM, on the other hand, is an optimization method built for the same setting as ours. We summarize the results in Table~\ref{tab:tab_amass_completion}. 
As can be seen, our method outperforms the two baselines by a large margin in all reported metrics. Note that on some of the examples (about $30\%$) FARM crashed during the optimization. We therefore only report the errors on its successful runs. Visualizations of several completions are shown in Figure~\ref{fig:amass_completion}. Additional completions are visualized in Appendix \ref{additional_vis}.

\input{tables/table_range_scan_completion.tex}  
\input{tables/table_amass_completion.tex}

\subsection{Non-rigid partial correspondences}

Finding dense correspondences between a full shape and its deformed parts is still an active research topic. Here we propose a solution in the form of alignment between the full shape and the partial shape, allowing for the recovery of the correspondence by a simple nearest neighbor search. As before, we evaluate this task on both FAUST and AMASS. 

\paragraph{FAUST projections}

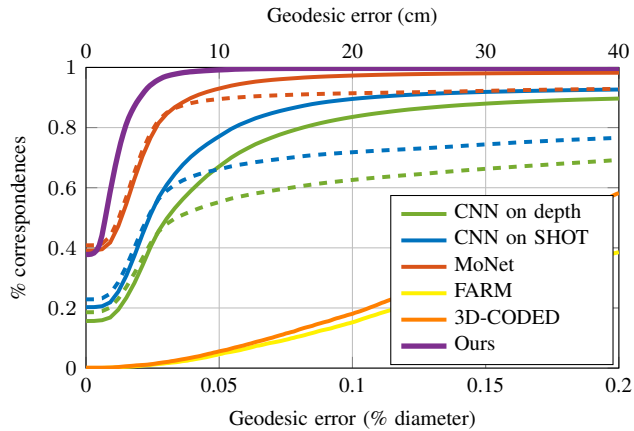
\begin{figure}[htbp]
	\centering
	\setlength\figureheight{4cm}
	\setlength\figurewidth{\linewidth}
	\input{./figures/Faust_Geoerr.tikz}
\caption{\textbf{Partial correspondence error curves, FAUST dataset}. Dashed line and solid line in the same color indicate performance before and after refinement, respectively. Note that our method doesn't require refinement, contributing to its computational speed.}
\label{fig:faust_corr_curve}
\end{figure}

On the FAUST projections dataset, we compare with two alignment-based methods, FARM and 3D-CODED. We also compare with 3 methods designed to only recover correspondences, i.e. without performing shape completion: MoNet~\cite{monet}, and two 3-layered Euclidean CNN baselines, trained on either SHOT~\cite{SHOT} descriptors or depth maps. Results are reported in Figure \ref{fig:faust_corr_curve}. As in the single view completion experiment, the test set consists of 200 shapes: 2 subjects at 10 different poses and 10 projected views. The direct matching baselines solve a labeling problem, assigning each input vertex a matching index in a fixed template shape. Differently, 3D-CODED deforms a fixed template and recovers correspondence by a nearest neighbor query for each input vertex using a one-sided Chamfer distance, as suggested in \cite{groueix20183d}. Our method and FARM both require a complete shape as input, which we chose as the null pose of each of the test examples. Due to slow convergence and unstable behavior of FARM we only kept 20 useful matching results on which we report the performance. As seen in Figure \ref{fig:faust_corr_curve}, our method outperforms prior art by a significant margin. This result is particularly interesting since it demonstrates that even though we solve an alignment problem, which is strictly harder than correspondence, we receive better results than methods that specialize in the latter. At the same time, looking at the poor performance demonstrated by the other alignment-based methods, we conclude that simply solving an alignment problem is not enough and the details of our method and training scheme allow for a substantial difference. Qualitative correspondence results are visualized in Appendix \ref{dense_corr}.

\begin{figure}[htbp]
    \centering
    \includegraphics[width=0.62\linewidth]{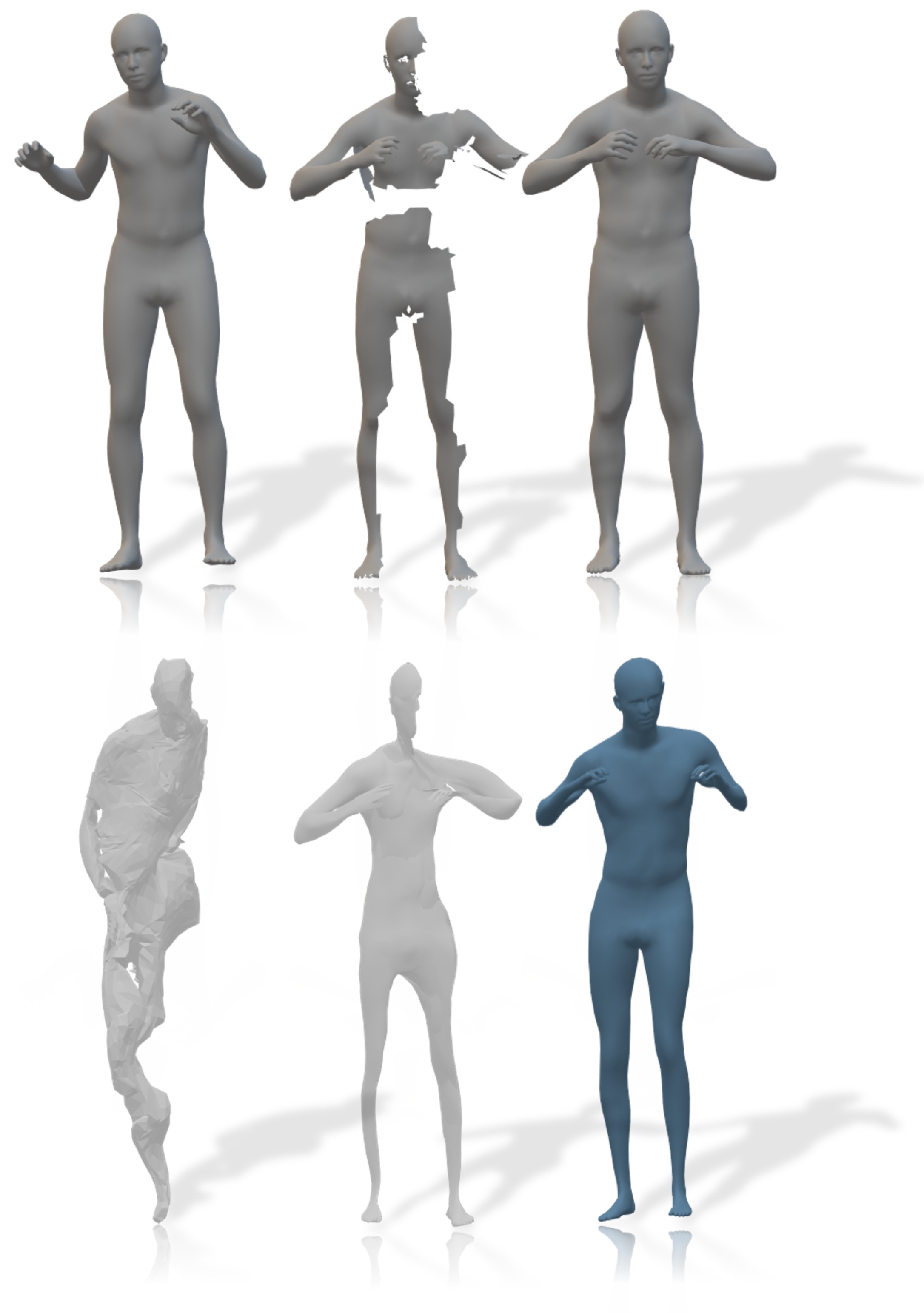}
    \caption{\textbf{AMASS Shape Completion}. At the top from left to right: full shape $Q$, partial shape $P$, ground truth completion $R$. At the bottom from left to right: reconstructions of FARM \cite{marin2018farm}, 3D-CODED \cite{groueix20183d} and ours.}
    \label{fig:amass_completion}
\end{figure}

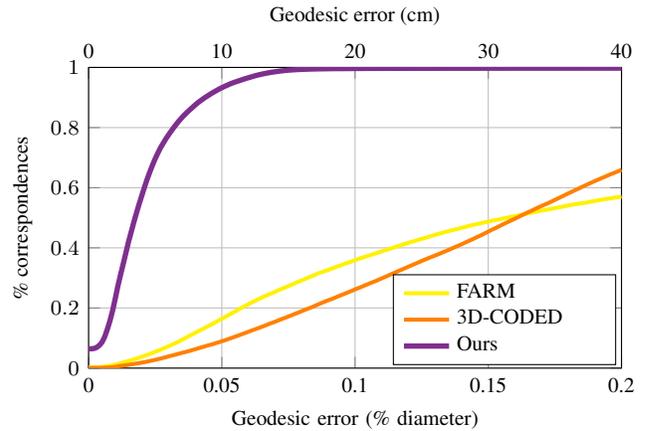
\begin{figure}[htbp]
	\centering
	\setlength\figureheight{4cm}
	\setlength\figurewidth{\linewidth}
	\input{./figures/Amass_Geoerr.tikz}
\caption{\textbf{Partial correspondence error curves, AMASS dataset}}
\label{fig:amass_corr_curve}
\end{figure}

\paragraph{AMASS Projections}

As FAUST is limited in variability, we further test our method on the recently published AMASS dataset. On the task of partial correspondence, we compare with FARM \cite{marin2018farm} and 3D-CODED \cite{groueix20183d} for which code was available online. We report the correspondence error graphs in Figure \ref{fig:amass_corr_curve}.
For evaluation we used 200 pairs of partial and full shapes chosen randomly (but consistently between different methods). Specifically, for each of the $4$ subjects in AMASS test set we randomized $50$ pairs of full poses: one was taken as the full shape $Q$ and one was projected to obtain the partial shape $P$, using the full unprojected version as the ground truth completion $R$. As with FAUST, we report the error curve of FARM taking the average of only the successful runs. As can be observed, our method outperforms both methods by a large margin. Qualitative correspondence results are visualized in Appendix \ref{dense_corr}.

\subsection{Real scans}

\begin{figure}[t!]
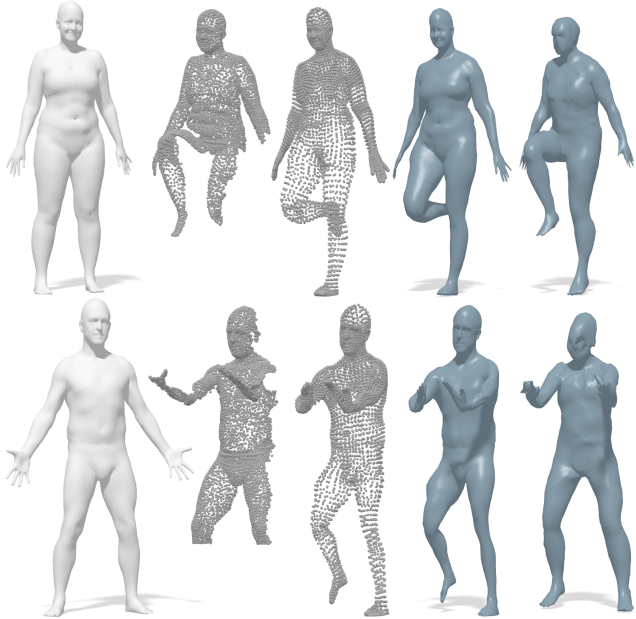

    \centering
    \includegraphics[width=\linewidth]{figures/knees_render.png}
    \includegraphics[width=\linewidth]{figures/arms_render.png}
    \caption{\textbf{Completion from real scans from the Dynamic Faust dataset}~\cite{dfaust:CVPR:2017}. 
    From left to right: Input reference shape; input raw scan; our completed shape as a point cloud; and as mesh; completion from Litany \etal ~\cite{litany2017deformable}. }
    \label{fig:real_dfaust_completion}
\end{figure}

To evaluate our method in real-world conditions, we test it on raw measurements taken during the preparation of the Dynamic FAUST~\cite{dfaust:CVPR:2017} dataset. This use case nicely matches our setting: these are partial scans of a subject for which we have a complete reference shape at a different pose. As pre-processing we compute point normals for the input scan using the method presented in \cite{hoppe1992surface}. The point cloud and the reference shape are subsequently inserted into a network pre-trained on FAUST. 
The template, raw scan, and our reconstruction are shown, from left to right, in Figure~\ref{fig:real_dfaust_completion}. We show our result both as the recovered point cloud as well as the recovered mesh using the template triangulation. As apparent from the figure, this is a challenging test case as it introduces several properties not seen at test time: a point cloud without connectivity leads to noisier normals, scanner noise, different point density and extreme partiality (note the missing bottom half of the shapes). Despite all these, the proposed network was able to recover the input quite elegantly, preserving shape details and mimicking the desired pose. In the rightmost column, we report a comparison with Litany \etal~\cite{litany2017deformable}. 
Note that while \cite{litany2017deformable} was trained on Dynamic FAUST, our network was trained on FAUST which is severely constrained in its pose variability. The result highlights that our method captures appearance details while pose accuracy is limited by the variability of the training set. 

%% file: tables/table_range_scan_completion.tex
\begin{table*}[t!]
\centering
\small

\begin{tabular}{ l@{\hskip 0.01\textwidth}c@{\hskip 0.01\textwidth}c@{\hskip 0.01\textwidth}c@{\hskip 0.01\textwidth}c@{\hskip 0.01\textwidth}c@{\hskip 0.01\textwidth}c@{\hskip 0.01\textwidth}c  }
    \hline\hline
    Error & Euclidean distance& Volumetric err. & Directional Chamfer distance & Directional Chamfer distance & Full Chamfer\\     
          & GT and reconstruction [cm] & mean $\pm$ std [\%] &   GT to reconstruction [cm]& reconstruction to GT [cm] & distance [cm]\\ \hline    
          
    Poisson \cite{kazhdan2013screened}       &   $-$        &   $24.8 \pm 23.2$   &   $7.3$     &   $3.64$        &    $10.94$ \\
    3D-EPN \cite{dai2016shape}               &   $-$        &   $89.7 \pm 33.8$   &   $4.52$    &  $4.87$    &  $9.39$ \\
    3D-CODED \cite{groueix20183d}      &   $35.50$  &   $21.8 \pm 0.3$           &   $11.15$   &   $38.49$  &  $49.64$ \\
    FARM \cite{marin2018farm}          &   $35.77$  &   $43.08 \pm 20.4$           &   $9.5$     &   $3.9$    &   $13.4$ \\
    Litany \etal~\cite{litany2017deformable} &   $7.07$        &   $9.24\pm8.62$     &   $2.84$    &   $2.9$        &   $5.74$\\

    {\bf Ours}  & $\textbf{2.94}$ 	& $\textbf{7.05} \pm \textbf{3.45}$           &  $\textbf{2.42}$ &  $\textbf{1.95}$   & $\textbf{4.37}$ \\ \hline \hline
    

  \end{tabular}
  \vspace{2mm}

\caption{\small \textbf{FAUST Shape Completion}. Comparison of different methods with respect to errors in vertex position and shape volume.
}
\label{tab_proj_completion}
\end{table*}

%% file: tables/table_amass_completion.tex
\begin{table*}[t!]
\centering
\small
\begin{tabular}{ l@{\hskip 0.01\textwidth}c@{\hskip 0.01\textwidth}c@{\hskip 0.01\textwidth}c@{\hskip 0.01\textwidth}c@{\hskip 0.01\textwidth}c@{\hskip 0.01\textwidth}c@{\hskip 0.01\textwidth}c  }
    \hline\hline
    Error & Euclidean distance& Volumetric err. & Directional Chamfer distance & Directional Chamfer distance & Full Chamfer\\     
          & GT and reconstruction [cm] & mean $\pm$ std [\%] &   GT to reconstruction [cm]& reconstruction to GT [cm] & distance [cm]\\ \hline    
          
    3D-CODED \cite{groueix20183d}      &   $36.14$  &   $\mathbf{14.84} \pm \mathbf{8.02}$           &   $13.65$   &   $35.35$  &  $49$ \\
    FARM \cite{marin2018farm}          &   $27.75$  &   $49.42 \pm 29.12$           &   $11.17$     &   $5.14$    &   $16.31$ \\
    {\bf Ours}  & $\mathbf{6.58}$ 	& $27.62 \pm 15.27$           &  $\mathbf{4.86}$ &  $\mathbf{3.06}$   & $\mathbf{7.92}$ \\ \hline \hline

  \end{tabular}
  \vspace{2mm}
  \vspace{2mm}
\caption{\label{tab:tab_amass_completion}\small \textbf{AMASS Shape Completion}. Comparison of different methods with respect to errors in vertex position and shape volume.
}
\label{table_amass}
\end{table*}

%% file: figures/Faust_Geoerr.tikz
%
%
\definecolor{cool_yellow}{rgb}{0.92900,0.69400,0.12500}%
\definecolor{cool_green}{rgb}{0.46600,0.67400,0.18800}%
\definecolor{cool_purple}{rgb}{0.49400,0.18400,0.55600}%
\definecolor{cool_blue}{rgb}{0.00000,0.44700,0.74100}%
\definecolor{cool_red}{rgb}{0.85000,0.32500,0.09800}%
\begin{tikzpicture}

\pgfplotsset{compat=newest} 

\tikzstyle{every node}=[font=\footnotesize]

\begin{axis}[%
width=0.85\figurewidth,
height=\figureheight,
scale only axis,
xmin=0,
xmax=0.2,
xlabel={Geodesic error (cm)},
xtick = {0,0.05,0.1,0.15,0.2},
xticklabels = {0,10,20,30,40},
axis x line*=top,
axis y line=none
]
\addplot [color=white,opacity=0.0,line width=1pt,forget plot]
  table[row sep=crcr]{%
0	0.0661103047895501\\
0.203030303030303	0.998846153846154\\
};
\end{axis}

\begin{axis}[%
width=0.85\figurewidth,
height=\figureheight,
scale only axis,
xmin=0,
xmax=0.2,
xlabel style={align=center,text width=5cm},
xlabel={Geodesic error (\% diameter)},
xtick={0,0.05,0.1,0.15,0.2},
xticklabels = {0,0.05,0.1,0.15,0.2},
xmajorgrids,
ymin=0,
ymax=1,
ytick={0,0.2,0.4,0.6,0.8,1},
ylabel={\% correspondences},
ymajorgrids,
axis background/.style={fill=white},
legend style={
	at={(0.99,0.01)},
	anchor=south east,
	legend cell align=left,
	align=left,
	text width=5.50em,
	text height=1ex
}
]
\addplot [color=cool_green,dashed,line width=1.5pt,forget plot]
  table[row sep=crcr]{%
0					0.185725446699647\\
0.00303030303030303	0.186023534483548\\
0.00606060606060606	0.189638493150138\\
0.00909090909090909	0.201471886260931\\
0.0121212121212121	0.23047151919219\\
0.0151515151515152	0.270107981129208\\
0.0181818181818182	0.315787022958624\\
0.0212121212121212	0.365742175287139\\
0.0242424242424242	0.415027995761537\\
0.0272727272727273	0.454385044459658\\
0.0303030303030303	0.479532793860501\\
0.0333333333333333	0.497534913988875\\
0.0363636363636364	0.510050626600633\\
0.0393939393939394	0.521286075894181\\
0.0424242424242424	0.531463340854177\\
0.0454545454545455	0.540719400333426\\
0.0484848484848485	0.548470097404602\\
0.0515151515151515	0.556202191180208\\
0.0545454545454545	0.564024146568903\\
0.0575757575757576	0.570128034239839\\
0.0606060606060606	0.575709060493571\\
0.0636363636363636	0.581120486278972\\
0.0666666666666667	0.585444189264572\\
0.0696969696969697	0.590079301326962\\
0.0727272727272727	0.594337650725814\\
0.0757575757575758	0.598434828006944\\
0.0787878787878788	0.6025925146672\\
0.0818181818181818	0.606719801050293\\
0.0848484848484849	0.610543369271526\\
0.0878787878787879	0.614114101744186\\
0.0909090909090909	0.617381144366093\\
0.0939393939393939	0.620195162626119\\
0.096969696969697	0.623254722527116\\
0.1				0.625939385626026\\
0.103030303030303	0.628527816982334\\
0.106060606060606	0.631197553693922\\
0.109090909090909	0.633665779426305\\
0.112121212121212	0.635822151484795\\
0.115151515151515	0.638110809159905\\
0.118181818181818	0.640428441556416\\
0.121212121212121	0.642449264137374\\
0.124242424242424	0.6445082621346\\
0.127272727272727	0.646617086968198\\
0.13030303030303	0.649109124249953\\
0.133333333333333	0.651449764310059\\
0.136363636363636	0.653801867878839\\
0.139393939393939	0.655902135969305\\
0.142424242424242	0.658082898104307\\
0.145454545454545	0.659996842370061\\
0.148484848484848	0.661923126889113\\
0.151515151515152	0.663907816454367\\
0.154545454545455	0.66569392841468\\
0.157575757575758	0.667542763204817\\
0.160606060606061	0.669464706574604\\
0.163636363636364	0.671301981437313\\
0.166666666666667	0.673001443920982\\
0.16969696969697	0.674674613728159\\
0.172727272727273	0.676455932866735\\
0.175757575757576	0.678464392933158\\
0.178787878787879	0.680221708599969\\
0.181818181818182	0.681814403718875\\
0.184848484848485	0.683591750854109\\
0.187878787878788	0.685306396749882\\
0.190909090909091	0.687041574745482\\
0.193939393939394	0.688659348475518\\
0.196969696969697	0.690677044985735\\
0.2	0.692505568863024\\
0.203030303030303	0.694278201079789\\
};

\addplot [color=cool_green,solid,line width=1.5pt]
  table[row sep=crcr]{%
0					0.156629303302702\\
0.00303030303030303	0.156841172523664\\
0.00606060606060606	0.159351067789971\\
0.00909090909090909	0.168235174264766\\
0.0121212121212121	0.194305129897017\\
0.0151515151515152	0.233953438714977\\
0.0181818181818182	0.283126997916003\\
0.0212121212121212	0.341458317203338\\
0.0242424242424242	0.403272583050636\\
0.0272727272727273	0.456558921285075\\
0.0303030303030303	0.49740640772489\\
0.0333333333333333	0.531857342392133\\
0.0363636363636364	0.562223239761336\\
0.0393939393939394	0.589857325909677\\
0.0424242424242424	0.61554104786611\\
0.0454545454545455	0.639980816987272\\
0.0484848484848485	0.660831374019259\\
0.0515151515151515	0.680814511960516\\
0.0545454545454545	0.700240832752971\\
0.0575757575757576	0.716673516175917\\
0.0606060606060606	0.731059009005233\\
0.0636363636363636	0.744815045538152\\
0.0666666666666667	0.756124448267607\\
0.0696969696969697	0.766646321282415\\
0.0727272727272727	0.77658882114044\\
0.0757575757575758	0.785320201615849\\
0.0787878787878788	0.793469807793904\\
0.0818181818181818	0.801122412990222\\
0.0848484848484849	0.808136933346782\\
0.0878787878787879	0.814650486948754\\
0.0909090909090909	0.820478134195713\\
0.0939393939393939	0.825842524083597\\
0.096969696969697		0.830811622371031\\
0.1				0.835431400960934\\
0.103030303030303	0.839705841273257\\
0.106060606060606	0.843754917587587\\
0.109090909090909	0.847455419211424\\
0.112121212121212	0.850749732019667\\
0.115151515151515	0.854080833606504\\
0.118181818181818	0.857192674975633\\
0.121212121212121	0.860214646312404\\
0.124242424242424	0.862679732333864\\
0.127272727272727	0.865147225743058\\
0.13030303030303	0.867471414119363\\
0.133333333333333	0.869780463633922\\
0.136363636363636	0.871938029527299\\
0.139393939393939	0.873901270588536\\
0.142424242424242	0.875773971497207\\
0.145454545454545	0.877213983838797\\
0.148484848484848	0.878942226879951\\
0.151515151515152	0.880528995741889\\
0.154545454545455	0.881914657131294\\
0.157575757575758	0.883409567527048\\
0.160606060606061	0.884665377282366\\
0.163636363636364	0.885847723259458\\
0.166666666666667	0.887075100293029\\
0.16969696969697	0.888116563066514\\
0.172727272727273	0.889230669993993\\
0.175757575757576	0.890233163107215\\
0.178787878787879	0.891227040627131\\
0.181818181818182	0.892086116003749\\
0.184848484848485	0.892955007296296\\
0.187878787878788	0.893689067850339\\
0.190909090909091	0.89445518880885\\
0.193939393939394	0.895188953403851\\
0.196969696969697	0.895842289468306\\
0.2				0.896576044746477\\
0.203030303030303	0.897185117995532\\
};
\addlegendentry{CNN on depth};

\addplot [color=cool_blue,dashed,line width=1.5pt,forget plot]
  table[row sep=crcr]{%
0					0.228650529722537\\
0.00303030303030303	0.228882340171733\\
0.00606060606060606	0.232261281215247\\
0.00909090909090909	0.247423636930395\\
0.0121212121212121	0.285998171762613\\
0.0151515151515152	0.338752284327589\\
0.0181818181818182	0.401194651135047\\
0.0212121212121212	0.46463397784349\\
0.0242424242424242	0.518377632797297\\
0.0272727272727273	0.562696422304376\\
0.0303030303030303	0.589983323158715\\
0.0333333333333333	0.610159519145129\\
0.0363636363636364	0.624556988474654\\
0.0393939393939394	0.636224571926929\\
0.0424242424242424	0.645126662409784\\
0.0454545454545455	0.65242498863426\\
0.0484848484848485	0.658695997482606\\
0.0515151515151515	0.664656279378075\\
0.0545454545454545	0.67116830551806\\
0.0575757575757576	0.67639932642116\\
0.0606060606060606	0.680614609897072\\
0.0636363636363636	0.68516543627888\\
0.0666666666666667	0.688707523372828\\
0.0696969696969697	0.692276327536462\\
0.0727272727272727	0.695502974039\\
0.0757575757575758	0.69890286782941\\
0.0787878787878788	0.7019961709243\\
0.0818181818181818	0.705306575144734\\
0.0848484848484849	0.708110188745514\\
0.0878787878787879	0.710319393633857\\
0.0909090909090909	0.712528127284598\\
0.0939393939393939	0.714355191180753\\
0.096969696969697		0.716377708830381\\
0.1					0.718037025915348\\
0.103030303030303	0.719677050130342\\
0.106060606060606	0.721403969579659\\
0.109090909090909	0.723020403964382\\
0.112121212121212	0.72446521902143\\
0.115151515151515	0.725841473587682\\
0.118181818181818	0.727278775641157\\
0.121212121212121	0.728639339969153\\
0.124242424242424	0.729939290662867\\
0.127272727272727	0.731436290618279\\
0.13030303030303	0.7330393805934\\
0.133333333333333	0.734709451832719\\
0.136363636363636	0.736277704553004\\
0.139393939393939	0.737961980220355\\
0.142424242424242	0.739815829928782\\
0.145454545454545	0.741515269905422\\
0.148484848484848	0.743325099550894\\
0.151515151515152	0.745064466687977\\
0.154545454545455	0.74642671507114\\
0.157575757575758	0.748195725914987\\
0.160606060606061	0.749933800507572\\
0.163636363636364	0.751550189226789\\
0.166666666666667	0.752815273296874\\
0.16969696969697	0.754141349139704\\
0.172727272727273	0.75537452710852\\
0.175757575757576	0.756914106575898\\
0.178787878787879	0.758181295285768\\
0.181818181818182	0.75937861252409\\
0.184848484848485	0.760514935881006\\
0.187878787878788	0.761575170608586\\
0.190909090909091	0.762768016199758\\
0.193939393939394	0.763828695950609\\
0.196969696969697	0.764926833145571\\
0.2				0.766014587560491\\
0.203030303030303	0.767092489749706\\
};

\addplot [color=cool_blue,solid,line width=1.5pt]
  table[row sep=crcr]{%
0					0.202764671966876\\
0.00303030303030303	0.20295641357158\\
0.00606060606060606	0.205853343226635\\
0.00909090909090909	0.218687466967991\\
0.0121212121212121	0.254161929500361\\
0.0151515151515152	0.307711715074547\\
0.0181818181818182	0.373950420122224\\
0.0212121212121212	0.445395493664996\\
0.0242424242424242	0.510845554811956\\
0.0272727272727273	0.567898572937444\\
0.0303030303030303	0.609801341346185\\
0.0333333333333333	0.645532228517175\\
0.0363636363636364	0.676522908513761\\
0.0393939393939394	0.703010514904831\\
0.0424242424242424	0.725218983755028\\
0.0454545454545455	0.74480343049978\\
0.0484848484848485	0.76252389331972\\
0.0515151515151515	0.779194219399228\\
0.0545454545454545	0.795802066409605\\
0.0575757575757576	0.809249769579802\\
0.0606060606060606	0.820231904363687\\
0.0636363636363636	0.831054251646452\\
0.0666666666666667	0.839784135995788\\
0.0696969696969697	0.847837061805634\\
0.0727272727272727	0.854998321657015\\
0.0757575757575758	0.861364677160199\\
0.0787878787878788	0.867287603505379\\
0.0818181818181818	0.872942608945379\\
0.0848484848484849	0.878118686124356\\
0.0878787878787879	0.882362104412984\\
0.0909090909090909	0.886135899956938\\
0.0939393939393939	0.889306208893042\\
0.096969696969697		0.892405218408884\\
0.1				0.895336071394115\\
0.103030303030303	0.897827264077693\\
0.106060606060606	0.900131290317517\\
0.109090909090909	0.90219838719928\\
0.112121212121212	0.904122368916717\\
0.115151515151515	0.905845579620551\\
0.118181818181818	0.907524953303712\\
0.121212121212121	0.908984895221549\\
0.124242424242424	0.910301134456252\\
0.127272727272727	0.911486791265089\\
0.13030303030303	0.912765423582197\\
0.133333333333333	0.913833633822263\\
0.136363636363636	0.914884163395035\\
0.139393939393939	0.915748464588389\\
0.142424242424242	0.916653131852404\\
0.145454545454545	0.917513063867616\\
0.148484848484848	0.918267543895559\\
0.151515151515152	0.919022364201326\\
0.154545454545455	0.919644114768338\\
0.157575757575758	0.920223035686353\\
0.160606060606061	0.920847997613097\\
0.163636363636364	0.921495315352184\\
0.166666666666667	0.922063895255541\\
0.16969696969697	0.922570471878475\\
0.172727272727273	0.923041419382175\\
0.175757575757576	0.923545020299706\\
0.178787878787879	0.924026951118277\\
0.181818181818182	0.924460377908688\\
0.184848484848485	0.924861205748305\\
0.187878787878788	0.925251556191959\\
0.190909090909091	0.925635102625503\\
0.193939393939394	0.925995906112658\\
0.196969696969697	0.92627720899\\
0.2				0.926603859494878\\
0.203030303030303	0.926887842244376\\
};
\addlegendentry{CNN on SHOT};

\addplot [color=cool_red,dashed,line width=1.5pt,forget plot]
  table[row sep=crcr]{%
0	0.408370627719179\\
0.00303030303030303	0.408657267347428\\
0.00606060606060606	0.413666943518781\\
0.00909090909090909	0.437892311232086\\
0.0121212121212121	0.494544772022521\\
0.0151515151515152	0.569252533764085\\
0.0181818181818182	0.651505019894929\\
0.0212121212121212	0.724446581588958\\
0.0242424242424242	0.782101706440681\\
0.0272727272727273	0.825617791236501\\
0.0303030303030303	0.849579694913194\\
0.0333333333333333	0.864680613117821\\
0.0363636363636364	0.873869613534966\\
0.0393939393939394	0.881404820843455\\
0.0424242424242424	0.886783309214408\\
0.0454545454545455	0.890541030215261\\
0.0484848484848485	0.893615324399433\\
0.0515151515151515	0.896152761566898\\
0.0545454545454545	0.898719126416659\\
0.0575757575757576	0.90068746217355\\
0.0606060606060606	0.902493957069216\\
0.0636363636363636	0.904179858574256\\
0.0666666666666667	0.905430528864379\\
0.0696969696969697	0.90662149472682\\
0.0727272727272727	0.907609241371448\\
0.0757575757575758	0.908657753939788\\
0.0787878787878788	0.909603852401356\\
0.0818181818181818	0.910520471463346\\
0.0848484848484849	0.911301820897791\\
0.0878787878787879	0.911937216650909\\
0.0909090909090909	0.912580374986417\\
0.0939393939393939	0.913176469978935\\
0.096969696969697	0.913754661832543\\
0.1	0.914270041143274\\
0.103030303030303	0.914746598188425\\
0.106060606060606	0.915158416201031\\
0.109090909090909	0.915611453340045\\
0.112121212121212	0.916004254450333\\
0.115151515151515	0.916475133448865\\
0.118181818181818	0.916915505542587\\
0.121212121212121	0.917307654927924\\
0.124242424242424	0.917701918493575\\
0.127272727272727	0.918159332417137\\
0.13030303030303	0.918676529606055\\
0.133333333333333	0.91917139563778\\
0.136363636363636	0.919689955425372\\
0.139393939393939	0.920168706446499\\
0.142424242424242	0.920819423151516\\
0.145454545454545	0.921409266302896\\
0.148484848484848	0.921972045098421\\
0.151515151515152	0.922512002075549\\
0.154545454545455	0.92305380063747\\
0.157575757575758	0.923607264288809\\
0.160606060606061	0.924139926353199\\
0.163636363636364	0.924677409333945\\
0.166666666666667	0.925101256860754\\
0.16969696969697	0.925561440031671\\
0.172727272727273	0.926042257960151\\
0.175757575757576	0.92645827578292\\
0.178787878787879	0.926852024338775\\
0.181818181818182	0.927127098248151\\
0.184848484848485	0.927378116482632\\
0.187878787878788	0.92764578362053\\
0.190909090909091	0.927911582893511\\
0.193939393939394	0.928170579628771\\
0.196969696969697	0.928410450749994\\
0.2	0.928612485838958\\
0.203030303030303	0.928873783478433\\
};

\addplot [color=cool_red,solid,line width=1.5pt]
  table[row sep=crcr]{%
0	0.390889533149985\\
0.00303030303030303	0.391185207136256\\
0.00606060606060606	0.39549392522673\\
0.00909090909090909	0.417761386844243\\
0.0121212121212121	0.472157179447104\\
0.0151515151515152	0.54636312209765\\
0.0181818181818182	0.629351581020332\\
0.0212121212121212	0.704894498654206\\
0.0242424242424242	0.766604272336185\\
0.0272727272727273	0.815349916297488\\
0.0303030303030303	0.845792951149432\\
0.0333333333333333	0.867880110525343\\
0.0363636363636364	0.884188904866235\\
0.0393939393939394	0.898580465069125\\
0.0424242424242424	0.909793928852119\\
0.0454545454545455	0.918830704441524\\
0.0484848484848485	0.926526918601316\\
0.0515151515151515	0.933139842617017\\
0.0545454545454545	0.939530596580261\\
0.0575757575757576	0.94464675892047\\
0.0606060606060606	0.948932913595948\\
0.0636363636363636	0.952916311419345\\
0.0666666666666667	0.955914967804957\\
0.0696969696969697	0.958610583249153\\
0.0727272727272727	0.960825207899753\\
0.0757575757575758	0.962736895597603\\
0.0787878787878788	0.964569551441538\\
0.0818181818181818	0.966499880240762\\
0.0848484848484849	0.967982387995336\\
0.0878787878787879	0.969229807375908\\
0.0909090909090909	0.970361475966655\\
0.0939393939393939	0.971288024940208\\
0.096969696969697	0.972256799673878\\
0.1	0.973140688663762\\
0.103030303030303	0.973951021265374\\
0.106060606060606	0.974638778097915\\
0.109090909090909	0.975346935647354\\
0.112121212121212	0.975960254172196\\
0.115151515151515	0.976514070900013\\
0.118181818181818	0.977067131293172\\
0.121212121212121	0.97754155017452\\
0.124242424242424	0.977922790275951\\
0.127272727272727	0.978315320092399\\
0.13030303030303	0.978631924201\\
0.133333333333333	0.979002589712218\\
0.136363636363636	0.979316540300662\\
0.139393939393939	0.979580095221407\\
0.142424242424242	0.979906064920623\\
0.145454545454545	0.980136107848673\\
0.148484848484848	0.980338183740563\\
0.151515151515152	0.980518067238846\\
0.154545454545455	0.980747070068846\\
0.157575757575758	0.980919531962497\\
0.160606060606061	0.981058678253135\\
0.163636363636364	0.981229169068382\\
0.166666666666667	0.981382258075152\\
0.16969696969697	0.981504974471381\\
0.172727272727273	0.981642623018533\\
0.175757575757576	0.981765054618107\\
0.178787878787879	0.981853167692299\\
0.181818181818182	0.981941216367571\\
0.184848484848485	0.982022788762691\\
0.187878787878788	0.98211774965124\\
0.190909090909091	0.98220249009883\\
0.193939393939394	0.982277425672161\\
0.196969696969697	0.982361904415547\\
0.2	0.982421506070372\\
0.203030303030303	0.982497197276045\\
};
\addlegendentry{MoNet};

\addplot [color=yellow,solid,line width=1.50pt]
  table[row sep=crcr]{%
0	0.000670875120380942\\
0.00100000000000000	0.000670875120380942\\
0.00200000000000000	0.000670875120380942\\
0.00300000000000000	0.000670875120380942\\
0.00400000000000000	0.000670875120380942\\
0.00500000000000000	0.000670875120380942\\
0.00600000000000000	0.000744060132090544\\
0.00700000000000000	0.00111194859705475\\
0.00800000000000000	0.00148330947457423\\
0.00900000000000000	0.00164714417035348\\
0.0100000000000000	0.00186854420577748\\
0.0110000000000000	0.00232904576122753\\
0.0120000000000000	0.00256834280823719\\
0.0130000000000000	0.00331511261907418\\
0.0140000000000000	0.00370262468979825\\
0.0150000000000000	0.00437609108280463\\
0.0160000000000000	0.00474975136089643\\
0.0170000000000000	0.00535015114831421\\
0.0180000000000000	0.00588556430964806\\
0.0190000000000000	0.00673585527534300\\
0.0200000000000000	0.00742199761215406\\
0.0210000000000000	0.00831592224127059\\
0.0220000000000000	0.00907734876850344\\
0.0230000000000000	0.00985449414971526\\
0.0240000000000000	0.0103800217584757\\
0.0250000000000000	0.0115451106449408\\
0.0260000000000000	0.0125473258591474\\
0.0270000000000000	0.0134722539248480\\
0.0280000000000000	0.0142193376687897\\
0.0290000000000000	0.0151128106603573\\
0.0300000000000000	0.0163387638330904\\
0.0310000000000000	0.0170230391085547\\
0.0320000000000000	0.0181441355862228\\
0.0330000000000000	0.0189671919703771\\
0.0340000000000000	0.0197984622986253\\
0.0350000000000000	0.0211115486207696\\
0.0360000000000000	0.0228443703470364\\
0.0370000000000000	0.0238160346669042\\
0.0380000000000000	0.0247250282227389\\
0.0390000000000000	0.0268715024471566\\
0.0400000000000000	0.0281388477052783\\
0.0410000000000000	0.0298264589839657\\
0.0420000000000000	0.0314670302488054\\
0.0430000000000000	0.0329170710746573\\
0.0440000000000000	0.0345870964380191\\
0.0450000000000000	0.0360217736282551\\
0.0460000000000000	0.0380598356935437\\
0.0470000000000000	0.0401183928767000\\
0.0480000000000000	0.0422393473454630\\
0.0490000000000000	0.0439698817767297\\
0.0500000000000000	0.0461817778016155\\
0.0510000000000000	0.0478419176865202\\
0.0520000000000000	0.0505187753344185\\
0.0530000000000000	0.0520426675607194\\
0.0540000000000000	0.0533432184062316\\
0.0550000000000000	0.0550915920265446\\
0.0560000000000000	0.0567125864300434\\
0.0570000000000000	0.0584428895728385\\
0.0580000000000000	0.0602045091806260\\
0.0590000000000000	0.0626559215741469\\
0.0600000000000000	0.0646100534967675\\
0.0610000000000000	0.0666991221470361\\
0.0620000000000000	0.0689120590790335\\
0.0630000000000000	0.0713446501522714\\
0.0640000000000000	0.0741569679515101\\
0.0650000000000000	0.0751293950051867\\
0.0660000000000000	0.0769380040922228\\
0.0670000000000000	0.0792345852237953\\
0.0680000000000000	0.0814999819969329\\
0.0690000000000000	0.0833781263369240\\
0.0700000000000000	0.0849629791945623\\
0.0710000000000000	0.0865268359027654\\
0.0720000000000000	0.0882732836117404\\
0.0730000000000000	0.0900036004549720\\
0.0740000000000000	0.0921687409273674\\
0.0750000000000000	0.0942890261705464\\
0.0760000000000000	0.0962444945380704\\
0.0770000000000000	0.0990774892817239\\
0.0780000000000000	0.100733266274716\\
0.0790000000000000	0.103473367824118\\
0.0800000000000000	0.105231547943994\\
0.0810000000000000	0.107307275995309\\
0.0820000000000000	0.109509132558655\\
0.0830000000000000	0.112199865346332\\
0.0840000000000000	0.114584985926215\\
0.0850000000000000	0.116854976805620\\
0.0860000000000000	0.119287190402572\\
0.0870000000000000	0.122040649842987\\
0.0880000000000000	0.124550826698010\\
0.0890000000000000	0.126581046707364\\
0.0900000000000000	0.128457120238822\\
0.0910000000000000	0.130302886880935\\
0.0920000000000000	0.132312253264616\\
0.0930000000000000	0.135181317442015\\
0.0940000000000000	0.137805108747237\\
0.0950000000000000	0.140040330737586\\
0.0960000000000000	0.142603643277230\\
0.0970000000000000	0.144969516312578\\
0.0980000000000000	0.147749799931918\\
0.0990000000000000	0.149645422536235\\
0.100000000000000	0.152278467787966\\
0.101000000000000	0.154815501415484\\
0.102000000000000	0.157916991780904\\
0.103000000000000	0.161084598202787\\
0.104000000000000	0.164245779011911\\
0.105000000000000	0.167031236954613\\
0.106000000000000	0.169992710756060\\
0.107000000000000	0.172853749456113\\
0.108000000000000	0.175302432226511\\
0.109000000000000	0.178101760283042\\
0.110000000000000	0.181352501099054\\
0.111000000000000	0.184150789227286\\
0.112000000000000	0.186739611562003\\
0.113000000000000	0.189461417515818\\
0.114000000000000	0.192080473146031\\
0.115000000000000	0.194466247055745\\
0.116000000000000	0.197322747489596\\
0.117000000000000	0.199950796142646\\
0.118000000000000	0.202932838940789\\
0.119000000000000	0.206292178889951\\
0.120000000000000	0.208972807583684\\
0.121000000000000	0.212493349546300\\
0.122000000000000	0.214767112374551\\
0.123000000000000	0.216717676935853\\
0.124000000000000	0.219307714546624\\
0.125000000000000	0.221829169811127\\
0.126000000000000	0.224458043715402\\
0.127000000000000	0.227170830676917\\
0.128000000000000	0.229616159059483\\
0.129000000000000	0.231204652664105\\
0.130000000000000	0.233321008134851\\
0.131000000000000	0.235753439070955\\
0.132000000000000	0.238154963944682\\
0.133000000000000	0.240660774565108\\
0.134000000000000	0.243272309470010\\
0.135000000000000	0.245617364516732\\
0.136000000000000	0.248040632048108\\
0.137000000000000	0.249774153302286\\
0.138000000000000	0.252382110110136\\
0.139000000000000	0.254427557784253\\
0.140000000000000	0.257420119524949\\
0.141000000000000	0.259150947561859\\
0.142000000000000	0.261536389092830\\
0.143000000000000	0.263148007279402\\
0.144000000000000	0.266020167061428\\
0.145000000000000	0.268136127762143\\
0.146000000000000	0.270821726702210\\
0.147000000000000	0.273115318831526\\
0.148000000000000	0.275671007673765\\
0.149000000000000	0.278304111865575\\
0.150000000000000	0.280385368955692\\
0.151000000000000	0.282280956047404\\
0.152000000000000	0.284931963538853\\
0.153000000000000	0.287987196166338\\
0.154000000000000	0.289939185494282\\
0.155000000000000	0.292868311098849\\
0.156000000000000	0.294607319127758\\
0.157000000000000	0.296879554678998\\
0.158000000000000	0.299207089665891\\
0.159000000000000	0.300980986707597\\
0.160000000000000	0.303741505092352\\
0.161000000000000	0.305633792208303\\
0.162000000000000	0.307843688037832\\
0.163000000000000	0.309749965605881\\
0.164000000000000	0.312017958222079\\
0.165000000000000	0.314087049781815\\
0.166000000000000	0.316547638047647\\
0.167000000000000	0.319784821778239\\
0.168000000000000	0.321634163680559\\
0.169000000000000	0.323723491079679\\
0.170000000000000	0.325513247442537\\
0.171000000000000	0.327538843043296\\
0.172000000000000	0.330316940716351\\
0.173000000000000	0.332445717047789\\
0.174000000000000	0.334403225868475\\
0.175000000000000	0.336404790393943\\
0.176000000000000	0.338687145216197\\
0.177000000000000	0.340969831213289\\
0.178000000000000	0.342551407111240\\
0.179000000000000	0.344236760274613\\
0.180000000000000	0.346580528270450\\
0.181000000000000	0.348253225693038\\
0.182000000000000	0.349953018963585\\
0.183000000000000	0.351684346287160\\
0.184000000000000	0.354234069845878\\
0.185000000000000	0.356292784737940\\
0.186000000000000	0.358869074555625\\
0.187000000000000	0.360598100883763\\
0.188000000000000	0.362516679048351\\
0.189000000000000	0.364888306604767\\
0.190000000000000	0.366764601464115\\
0.191000000000000	0.368134840915582\\
0.192000000000000	0.370805082200805\\
0.193000000000000	0.372891249386885\\
0.194000000000000	0.374546052436174\\
0.195000000000000	0.376823222026251\\
0.196000000000000	0.378042537679954\\
0.197000000000000	0.379481771356938\\
0.198000000000000	0.381486719812686\\
0.199000000000000	0.383578737666347\\
0.200000000000000	0.385718729111279\\
};
\addlegendentry{FARM};

\addplot [color=orange,solid,line width=1.5pt]
  table[row sep=crcr]{%
0	0.00142192264629520\\ 
0.00100000000000000	0.00142192264629520\\ 
0.00200000000000000	0.00142192264629520\\ 
0.00300000000000000	0.00142192264629520\\ 
0.00400000000000000	0.00142192264629520\\ 
0.00500000000000000	0.00142192264629520\\ 
0.00600000000000000	0.00142192264629520\\ 
0.00700000000000000	0.00147390850614132\\ 
0.00800000000000000	0.00168877243885413\\ 
0.00900000000000000	0.00187112271001095\\ 
0.0100000000000000	0.00221888102378825\\ 
0.0110000000000000	0.00298223713466019\\ 
0.0120000000000000	0.00346529572472968\\ 
0.0130000000000000	0.00434238885675451\\ 
0.0140000000000000	0.00527703711933399\\ 
0.0150000000000000	0.00604150973281653\\ 
0.0160000000000000	0.00670612774382735\\ 
0.0170000000000000	0.00745063954699532\\ 
0.0180000000000000	0.00826848868332694\\ 
0.0190000000000000	0.00919892677774656\\ 
0.0200000000000000	0.00963394243695366\\ 
0.0210000000000000	0.0103195176612543\\ 
0.0220000000000000	0.0107813329092271\\ 
0.0230000000000000	0.0114866014275174\\ 
0.0240000000000000	0.0126653755798701\\ 
0.0250000000000000	0.0136056820431883\\ 
0.0260000000000000	0.0149878483036482\\ 
0.0270000000000000	0.0165492711440246\\ 
0.0280000000000000	0.0177368393730465\\ 
0.0290000000000000	0.0190020866623962\\ 
0.0300000000000000	0.0200579747007746\\ 
0.0310000000000000	0.0210980416129924\\ 
0.0320000000000000	0.0230233487213123\\ 
0.0330000000000000	0.0241603527115083\\ 
0.0340000000000000	0.0257219961596596\\ 
0.0350000000000000	0.0273923497279095\\ 
0.0360000000000000	0.0290230540883430\\ 
0.0370000000000000	0.0305007926264064\\ 
0.0380000000000000	0.0320354099009226\\ 
0.0390000000000000	0.0337557589200941\\ 
0.0400000000000000	0.0355418609445364\\ 
0.0410000000000000	0.0373407780086119\\ 
0.0420000000000000	0.0388526025289126\\ 
0.0430000000000000	0.0407297826002758\\ 
0.0440000000000000	0.0431411509627082\\ 
0.0450000000000000	0.0452286669057082\\ 
0.0460000000000000	0.0474579416251069\\ 
0.0470000000000000	0.0491679721293757\\ 
0.0480000000000000	0.0517538955403330\\ 
0.0490000000000000	0.0534342751153205\\ 
0.0500000000000000	0.0558177546298385\\ 
0.0510000000000000	0.0577227337842254\\ 
0.0520000000000000	0.0595052670671188\\ 
0.0530000000000000	0.0612025420097507\\ 
0.0540000000000000	0.0637604977021424\\ 
0.0550000000000000	0.0663411520718888\\ 
0.0560000000000000	0.0688076743089381\\ 
0.0570000000000000	0.0706744022715520\\ 
0.0580000000000000	0.0729380350469823\\ 
0.0590000000000000	0.0748973227912418\\ 
0.0600000000000000	0.0775687339200375\\ 
0.0610000000000000	0.0800163098501924\\ 
0.0620000000000000	0.0822148718475823\\ 
0.0630000000000000	0.0849877323601936\\ 
0.0640000000000000	0.0875793414913159\\ 
0.0650000000000000	0.0896509894453214\\ 
0.0660000000000000	0.0918240658180153\\ 
0.0670000000000000	0.0943149531449042\\ 
0.0680000000000000	0.0968861260649067\\ 
0.0690000000000000	0.0992545946433371\\ 
0.0700000000000000	0.101516689729697\\ 
0.0710000000000000	0.104490008446108\\ 
0.0720000000000000	0.107087293551362\\ 
0.0730000000000000	0.109656743877443\\ 
0.0740000000000000	0.111888164648677\\ 
0.0750000000000000	0.115202140782081\\ 
0.0760000000000000	0.117172409747955\\ 
0.0770000000000000	0.119801592034752\\ 
0.0780000000000000	0.121721761303344\\ 
0.0790000000000000	0.125026088009683\\ 
0.0800000000000000	0.128414043296477\\ 
0.0810000000000000	0.131173450281338\\ 
0.0820000000000000	0.133864301127993\\ 
0.0830000000000000	0.136204833549880\\ 
0.0840000000000000	0.138370047373305\\ 
0.0850000000000000	0.140699686936631\\ 
0.0860000000000000	0.144065659087309\\ 
0.0870000000000000	0.146816641618569\\ 
0.0880000000000000	0.149210997269243\\ 
0.0890000000000000	0.151840169753111\\ 
0.0900000000000000	0.154126811124494\\ 
0.0910000000000000	0.156103032622870\\ 
0.0920000000000000	0.159229171938403\\ 
0.0930000000000000	0.162654597578537\\ 
0.0940000000000000	0.165603883890746\\ 
0.0950000000000000	0.167767826641059\\ 
0.0960000000000000	0.170497730301940\\ 
0.0970000000000000	0.172901123184349\\ 
0.0980000000000000	0.175998054480110\\ 
0.0990000000000000	0.178407926929073\\ 
0.100000000000000	0.180520387241661\\ 
0.101000000000000	0.184218511559592\\ 
0.102000000000000	0.186681836811526\\ 
0.103000000000000	0.189463607549170\\ 
0.104000000000000	0.192663096524078\\ 
0.105000000000000	0.195813310485409\\ 
0.106000000000000	0.199352336662582\\ 
0.107000000000000	0.202880090117643\\ 
0.108000000000000	0.205975874975190\\ 
0.109000000000000	0.209351023784248\\ 
0.110000000000000	0.213130557266200\\ 
0.111000000000000	0.216960671525274\\ 
0.112000000000000	0.219798175240675\\ 
0.113000000000000	0.223465872371189\\ 
0.114000000000000	0.226660267432848\\ 
0.115000000000000	0.229695584619846\\ 
0.116000000000000	0.233100879596275\\ 
0.117000000000000	0.236917531606886\\ 
0.118000000000000	0.240172436021813\\ 
0.119000000000000	0.244161668392728\\ 
0.120000000000000	0.248428562119066\\ 
0.121000000000000	0.251832548153810\\ 
0.122000000000000	0.255419783936857\\ 
0.123000000000000	0.259038146667901\\ 
0.124000000000000	0.263531695514649\\ 
0.125000000000000	0.267373637726515\\ 
0.126000000000000	0.271669079713091\\ 
0.127000000000000	0.275707541958932\\ 
0.128000000000000	0.279102195060471\\ 
0.129000000000000	0.283685058936359\\ 
0.130000000000000	0.287598409409451\\ 
0.131000000000000	0.291562688677153\\ 
0.132000000000000	0.295220402432245\\ 
0.133000000000000	0.299203677003269\\ 
0.134000000000000	0.303600484999937\\ 
0.135000000000000	0.307376054504900\\ 
0.136000000000000	0.311995820546370\\ 
0.137000000000000	0.315473746428933\\ 
0.138000000000000	0.319380559862810\\ 
0.139000000000000	0.323237264044712\\ 
0.140000000000000	0.327025772776276\\ 
0.141000000000000	0.331855745773371\\ 
0.142000000000000	0.335744430012896\\ 
0.143000000000000	0.338860339571736\\ 
0.144000000000000	0.342704448012779\\ 
0.145000000000000	0.346072671428869\\ 
0.146000000000000	0.350280614371501\\ 
0.147000000000000	0.353858835963211\\ 
0.148000000000000	0.358337417500991\\ 
0.149000000000000	0.361526290035105\\ 
0.150000000000000	0.366315997373301\\ 
0.151000000000000	0.370940073538452\\ 
0.152000000000000	0.375810475452786\\ 
0.153000000000000	0.379840721230111\\ 
0.154000000000000	0.383429840131773\\ 
0.155000000000000	0.387267384528243\\ 
0.156000000000000	0.391511855623267\\ 
0.157000000000000	0.396135542997058\\ 
0.158000000000000	0.400653611222314\\ 
0.159000000000000	0.404782944381212\\ 
0.160000000000000	0.408812914675669\\ 
0.161000000000000	0.412906513637675\\ 
0.162000000000000	0.417472392608447\\ 
0.163000000000000	0.421701176785575\\ 
0.164000000000000	0.426384995822238\\ 
0.165000000000000	0.431192139581045\\ 
0.166000000000000	0.435307421423843\\ 
0.167000000000000	0.439739062353760\\ 
0.168000000000000	0.444218428681235\\ 
0.169000000000000	0.448585685519014\\ 
0.170000000000000	0.452460043976658\\ 
0.171000000000000	0.457694713975637\\ 
0.172000000000000	0.461577930133280\\ 
0.173000000000000	0.466110212350851\\ 
0.174000000000000	0.471016056487236\\ 
0.175000000000000	0.475451867130110\\ 
0.176000000000000	0.479654432457564\\ 
0.177000000000000	0.484795822128010\\ 
0.178000000000000	0.489487943664949\\ 
0.179000000000000	0.494710052589648\\ 
0.180000000000000	0.499475847963574\\ 
0.181000000000000	0.504910631648911\\ 
0.182000000000000	0.509292415889002\\ 
0.183000000000000	0.513455929327593\\ 
0.184000000000000	0.517726587299012\\ 
0.185000000000000	0.522704680536528\\ 
0.186000000000000	0.526769592170768\\ 
0.187000000000000	0.530485880228101\\ 
0.188000000000000	0.535330506223379\\ 
0.189000000000000	0.539026948421023\\ 
0.190000000000000	0.542764956093150\\ 
0.191000000000000	0.546700575638525\\ 
0.192000000000000	0.550564054567092\\ 
0.193000000000000	0.554852551864668\\ 
0.194000000000000	0.559173042418391\\ 
0.195000000000000	0.562925376428456\\ 
0.196000000000000	0.566073604978050\\ 
0.197000000000000	0.570301947989076\\ 
0.198000000000000	0.574763604237761\\ 
0.199000000000000	0.578861424009305\\ 
0.200000000000000	0.582249585939490\\
};
\addlegendentry{3D-CODED};

\addplot [color=cool_purple,solid,line width=2.0pt]
  table[row sep=crcr]{%
0.0	0.377951611998809\\
0.00100000000000000	0.378036033373903\\
0.00200000000000000	0.379167035155197\\
0.00300000000000000	0.382959244917291\\
0.00400000000000000	0.391535548561057\\
0.00500000000000000	0.408818636076324\\
0.00600000000000000	0.440863484994918\\
0.00700000000000000	0.486986791528920\\
0.00800000000000000	0.535073908425617\\
0.00900000000000000	0.583559115754113\\
0.0100000000000000	0.628341576956260\\
0.0110000000000000	0.672363368054937\\
0.0120000000000000	0.711595990977030\\
0.0130000000000000	0.748568452371760\\
0.0140000000000000	0.780159395085799\\
0.0150000000000000	0.808631424390558\\
0.0160000000000000	0.832118187383844\\
0.0170000000000000	0.851277095584864\\
0.0180000000000000	0.868261232595781\\
0.0190000000000000	0.883536241669241\\
0.0200000000000000	0.896601465316566\\
0.0210000000000000	0.908714600320218\\
0.0220000000000000	0.920553477737559\\
0.0230000000000000	0.931021438637458\\
0.0240000000000000	0.939638114344266\\
0.0250000000000000	0.946225505167257\\
0.0260000000000000	0.952889616305034\\
0.0270000000000000	0.958273704100909\\
0.0280000000000000	0.962960560438036\\
0.0290000000000000	0.967412618263656\\
0.0300000000000000	0.970573097325085\\
0.0310000000000000	0.972973217130731\\
0.0320000000000000	0.975261318147548\\
0.0330000000000000	0.977066744598635\\
0.0340000000000000	0.978870588760588\\
0.0350000000000000	0.980507384313727\\
0.0360000000000000	0.982060616490203\\
0.0370000000000000	0.983439454599471\\
0.0380000000000000	0.984334182555086\\
0.0390000000000000	0.985531718779241\\
0.0400000000000000	0.986060555942572\\
0.0410000000000000	0.986892094228688\\
0.0420000000000000	0.987611217343257\\
0.0430000000000000	0.988165374986298\\
0.0440000000000000	0.988850890393063\\
0.0450000000000000	0.989426205301674\\
0.0460000000000000	0.989884274719578\\
0.0470000000000000	0.990394589394571\\
0.0480000000000000	0.990940553344350\\
0.0490000000000000	0.991447205640969\\
0.0500000000000000	0.991881074568725\\
0.0510000000000000	0.992129842323015\\
0.0520000000000000	0.992563320439331\\
0.0530000000000000	0.992820604078365\\
0.0540000000000000	0.993358694946519\\
0.0550000000000000	0.993881761627720\\
0.0560000000000000	0.994384668270388\\
0.0570000000000000	0.994534066140127\\
0.0580000000000000	0.994825304733737\\
0.0590000000000000	0.995047304739619\\
0.0600000000000000	0.995316663843333\\
0.0610000000000000	0.995475507034166\\
0.0620000000000000	0.995648943162214\\
0.0630000000000000	0.995906392914526\\
0.0640000000000000	0.996073859127964\\
0.0650000000000000	0.996126268820448\\
0.0660000000000000	0.996238812226639\\
0.0670000000000000	0.996238812226639\\
0.0680000000000000	0.996332840307932\\
0.0690000000000000	0.996426517555212\\
0.0700000000000000	0.996487318217477\\
0.0710000000000000	0.996517080122239\\
0.0720000000000000	0.996569784122132\\
0.0730000000000000	0.996597271752699\\
0.0740000000000000	0.996665149206377\\
0.0750000000000000	0.996744528780451\\
0.0760000000000000	0.996771082152730\\
0.0770000000000000	0.996771082152730\\
0.0780000000000000	0.996771082152730\\
0.0790000000000000	0.996798569783296\\
0.0800000000000000	0.996798569783296\\
0.0810000000000000	0.996798569783296\\
0.0820000000000000	0.996824303185252\\
0.0830000000000000	0.996824303185252\\
0.0840000000000000	0.996824303185252\\
0.0850000000000000	0.996824303185252\\
0.0860000000000000	0.996824303185252\\
0.0870000000000000	0.996824303185252\\
0.0880000000000000	0.996824303185252\\
0.0890000000000000	0.996824303185252\\
0.0900000000000000	0.996824303185252\\
0.0910000000000000	0.996850856557530\\
0.0920000000000000	0.996850856557530\\
0.0930000000000000	0.996850856557530\\
0.0940000000000000	0.996850856557530\\
0.0950000000000000	0.996876589959485\\
0.0960000000000000	0.996876589959485\\
0.0970000000000000	0.996876589959485\\
0.0980000000000000	0.996876589959485\\
0.0990000000000000	0.996902323361441\\
0.100000000000000	0.996902323361441\\
0.101000000000000	0.996902323361441\\
0.102000000000000	0.996902323361441\\
0.103000000000000	0.996902323361441\\
0.104000000000000	0.996902323361441\\
0.105000000000000	0.996902323361441\\
0.106000000000000	0.996902323361441\\
0.107000000000000	0.996902323361441\\
0.108000000000000	0.996902323361441\\
0.109000000000000	0.996902323361441\\
0.110000000000000	0.996902323361441\\
0.111000000000000	0.996902323361441\\
0.112000000000000	0.996902323361441\\
0.113000000000000	0.996902323361441\\
0.114000000000000	0.996902323361441\\
0.115000000000000	0.996902323361441\\
0.116000000000000	0.996902323361441\\
0.117000000000000	0.996902323361441\\
0.118000000000000	0.996902323361441\\
0.119000000000000	0.996902323361441\\
0.120000000000000	0.996902323361441\\
0.121000000000000	0.996902323361441\\
0.122000000000000	0.996902323361441\\
0.123000000000000	0.996902323361441\\
0.124000000000000	0.996902323361441\\
0.125000000000000	0.996902323361441\\
0.126000000000000	0.996902323361441\\
0.127000000000000	0.996902323361441\\
0.128000000000000	0.996902323361441\\
0.129000000000000	0.996902323361441\\
0.130000000000000	0.996902323361441\\
0.131000000000000	0.996902323361441\\
0.132000000000000	0.996902323361441\\
0.133000000000000	0.996902323361441\\
0.134000000000000	0.996902323361441\\
0.135000000000000	0.996902323361441\\
0.136000000000000	0.996902323361441\\
0.137000000000000	0.996902323361441\\
0.138000000000000	0.996902323361441\\
0.139000000000000	0.996902323361441\\
0.140000000000000	0.996902323361441\\
0.141000000000000	0.996902323361441\\
0.142000000000000	0.996902323361441\\
0.143000000000000	0.996902323361441\\
0.144000000000000	0.996902323361441\\
0.145000000000000	0.996902323361441\\
0.146000000000000	0.996902323361441\\
0.147000000000000	0.996902323361441\\
0.148000000000000	0.996902323361441\\
0.149000000000000	0.996902323361441\\
0.150000000000000	0.996902323361441\\
0.151000000000000	0.996902323361441\\
0.152000000000000	0.996902323361441\\
0.153000000000000	0.996902323361441\\
0.154000000000000	0.996902323361441\\
0.155000000000000	0.996902323361441\\
0.156000000000000	0.996902323361441\\
0.157000000000000	0.996902323361441\\
0.158000000000000	0.996902323361441\\
0.159000000000000	0.996902323361441\\
0.160000000000000	0.996902323361441\\
0.161000000000000	0.996902323361441\\
0.162000000000000	0.996902323361441\\
0.163000000000000	0.996902323361441\\
0.164000000000000	0.996902323361441\\
0.165000000000000	0.996902323361441\\
0.166000000000000	0.996902323361441\\
0.167000000000000	0.996902323361441\\
0.168000000000000	0.996902323361441\\
0.169000000000000	0.996902323361441\\
0.170000000000000	0.996902323361441\\
0.171000000000000	0.996902323361441\\
0.172000000000000	0.996902323361441\\
0.173000000000000	0.996902323361441\\
0.174000000000000	0.996902323361441\\
0.175000000000000	0.996902323361441\\
0.176000000000000	0.996902323361441\\
0.177000000000000	0.996902323361441\\
0.178000000000000	0.996902323361441\\
0.179000000000000	0.996902323361441\\
0.180000000000000	0.996902323361441\\
0.181000000000000	0.996902323361441\\
0.182000000000000	0.996902323361441\\
0.183000000000000	0.996902323361441\\
0.184000000000000	0.996902323361441\\
0.185000000000000	0.996902323361441\\
0.186000000000000	0.996902323361441\\
0.187000000000000	0.996902323361441\\
0.188000000000000	0.996902323361441\\
0.189000000000000	0.996902323361441\\
0.190000000000000	0.996902323361441\\
0.191000000000000	0.996902323361441\\
0.192000000000000	0.996902323361441\\
0.193000000000000	0.996902323361441\\
0.194000000000000	0.996902323361441\\
0.195000000000000	0.996902323361441\\
0.196000000000000	0.996902323361441\\
0.197000000000000	0.996902323361441\\
0.198000000000000	0.996929810992008\\
0.199000000000000	0.996929810992008\\
0.200000000000000	0.996929810992008\\
};
\addlegendentry{Ours};

\end{axis}
\end{tikzpicture}

%% file: figures/Amass_Geoerr.tikz
%
%
\definecolor{cool_yellow}{rgb}{0.92900,0.69400,0.12500}%
\definecolor{cool_green}{rgb}{0.46600,0.67400,0.18800}%
\definecolor{cool_purple}{rgb}{0.49400,0.18400,0.55600}%
\definecolor{cool_blue}{rgb}{0.00000,0.44700,0.74100}%
\definecolor{cool_red}{rgb}{0.85000,0.32500,0.09800}%
\begin{tikzpicture}

\pgfplotsset{compat=newest} 

\tikzstyle{every node}=[font=\footnotesize]

\begin{axis}[%
width=0.85\figurewidth,
height=\figureheight,
scale only axis,
xmin=0,
xmax=0.2,
xlabel={Geodesic error (cm)},
xtick = {0,0.05,0.1,0.15,0.2},
xticklabels = {0,10,20,30,40},
axis x line*=top,
axis y line=none
]
\addplot [color=white,opacity=0.0,line width=1pt,forget plot]
  table[row sep=crcr]{%
0	0.0661103047895501\\
0.203030303030303	0.998846153846154\\
};
\end{axis}

\begin{axis}[%
width=0.85\figurewidth,
height=\figureheight,
scale only axis,
xmin=0,
xmax=0.2,
xlabel style={align=center,text width=5cm},
xlabel={Geodesic error (\% diameter)},
xtick={0,0.05,0.1,0.15,0.2},
xticklabels = {0,0.05,0.1,0.15,0.2},
xmajorgrids,
ymin=0,
ymax=1,
ytick={0,0.2,0.4,0.6,0.8,1},
ylabel={\% correspondences},
ymajorgrids,
axis background/.style={fill=white},
legend style={
	at={(0.99,0.01)},
	anchor=south east,
	legend cell align=left,
	align=left,
	text width=5.50em,
	text height=1ex
}
]

\addplot [color=yellow,solid,line width=1.50pt]
  table[row sep=crcr]{%
0	0.00240493457886378\\ 
0.00100000000000000	0.00240493457886378\\ 
0.00200000000000000	0.00243896115081208\\ 
0.00300000000000000	0.00274356798337005\\ 
0.00400000000000000	0.00322573806640401\\ 
0.00500000000000000	0.00394324700926630\\ 
0.00600000000000000	0.00510162192705341\\ 
0.00700000000000000	0.00632471240381658\\ 
0.00800000000000000	0.00788531547374805\\ 
0.00900000000000000	0.00973723018698097\\ 
0.0100000000000000	0.0113742446912645\\ 
0.0110000000000000	0.0136867979051790\\ 
0.0120000000000000	0.0159567983063069\\ 
0.0130000000000000	0.0188142363993049\\ 
0.0140000000000000	0.0213154710112945\\ 
0.0150000000000000	0.0239989314476856\\ 
0.0160000000000000	0.0264819332239405\\ 
0.0170000000000000	0.0292646408301330\\ 
0.0180000000000000	0.0319363948214103\\ 
0.0190000000000000	0.0349338113809077\\ 
0.0200000000000000	0.0378739360971826\\ 
0.0210000000000000	0.0408983979872810\\ 
0.0220000000000000	0.0442281180716323\\ 
0.0230000000000000	0.0471088735595896\\ 
0.0240000000000000	0.0506457705540608\\ 
0.0250000000000000	0.0538957044310398\\ 
0.0260000000000000	0.0578312968792915\\ 
0.0270000000000000	0.0611076764299702\\ 
0.0280000000000000	0.0651824366714874\\ 
0.0290000000000000	0.0691765521894802\\ 
0.0300000000000000	0.0730761365772482\\ 
0.0310000000000000	0.0765750012835045\\ 
0.0320000000000000	0.0807637283807967\\ 
0.0330000000000000	0.0851690596699943\\ 
0.0340000000000000	0.0893880551570509\\ 
0.0350000000000000	0.0939087878590015\\ 
0.0360000000000000	0.0983828440345134\\ 
0.0370000000000000	0.103159174759501\\ 
0.0380000000000000	0.107962173947069\\ 
0.0390000000000000	0.112644725161761\\ 
0.0400000000000000	0.117393428254784\\ 
0.0410000000000000	0.122061650586008\\ 
0.0420000000000000	0.126474737820349\\ 
0.0430000000000000	0.130916421497941\\ 
0.0440000000000000	0.135796827003058\\ 
0.0450000000000000	0.140161964596926\\ 
0.0460000000000000	0.144685749584407\\ 
0.0470000000000000	0.149501025436274\\ 
0.0480000000000000	0.154203189600005\\ 
0.0490000000000000	0.159658136162260\\ 
0.0500000000000000	0.164183744061895\\ 
0.0510000000000000	0.168919735703408\\ 
0.0520000000000000	0.173846569682685\\ 
0.0530000000000000	0.178785468939040\\ 
0.0540000000000000	0.183592529245044\\ 
0.0550000000000000	0.188406715653349\\ 
0.0560000000000000	0.193384387775841\\ 
0.0570000000000000	0.198736593879508\\ 
0.0580000000000000	0.203349982161154\\ 
0.0590000000000000	0.208056455488906\\ 
0.0600000000000000	0.212533617063513\\ 
0.0610000000000000	0.217414268075426\\ 
0.0620000000000000	0.221683199833420\\ 
0.0630000000000000	0.226318297097291\\ 
0.0640000000000000	0.230389755991298\\ 
0.0650000000000000	0.234723609309312\\ 
0.0660000000000000	0.238681647152113\\ 
0.0670000000000000	0.242484614372825\\ 
0.0680000000000000	0.246235087589066\\ 
0.0690000000000000	0.250014860861292\\ 
0.0700000000000000	0.253996007857085\\ 
0.0710000000000000	0.258237503406879\\ 
0.0720000000000000	0.262022886895371\\ 
0.0730000000000000	0.265679792541115\\ 
0.0740000000000000	0.269194024055055\\ 
0.0750000000000000	0.273114662836351\\ 
0.0760000000000000	0.276787877496805\\ 
0.0770000000000000	0.280470626543091\\ 
0.0780000000000000	0.284472823218081\\ 
0.0790000000000000	0.288431295550805\\ 
0.0800000000000000	0.291858471509637\\ 
0.0810000000000000	0.295657312532073\\ 
0.0820000000000000	0.299339631557309\\ 
0.0830000000000000	0.303311196056114\\ 
0.0840000000000000	0.306724277426688\\ 
0.0850000000000000	0.310433446536481\\ 
0.0860000000000000	0.314528392165794\\ 
0.0870000000000000	0.317740418332183\\ 
0.0880000000000000	0.320443350749277\\ 
0.0890000000000000	0.323658563967778\\ 
0.0900000000000000	0.327038913278354\\ 
0.0910000000000000	0.330247936031378\\ 
0.0920000000000000	0.333644343636716\\ 
0.0930000000000000	0.336991806707797\\ 
0.0940000000000000	0.340274334326330\\ 
0.0950000000000000	0.343252458080741\\ 
0.0960000000000000	0.346725951074033\\ 
0.0970000000000000	0.349678782078307\\ 
0.0980000000000000	0.352549510164466\\ 
0.0990000000000000	0.355831139633577\\ 
0.100000000000000	0.358983353470313\\ 
0.101000000000000	0.362114843908274\\ 
0.102000000000000	0.365444996140085\\ 
0.103000000000000	0.368022033501966\\ 
0.104000000000000	0.370931581484881\\ 
0.105000000000000	0.374033191824575\\ 
0.106000000000000	0.376381711780763\\ 
0.107000000000000	0.379523531492436\\ 
0.108000000000000	0.382534681425245\\ 
0.109000000000000	0.385528810849768\\ 
0.110000000000000	0.388394574284846\\ 
0.111000000000000	0.390968441543157\\ 
0.112000000000000	0.393750524064064\\ 
0.113000000000000	0.396307508496636\\ 
0.114000000000000	0.399661580929457\\ 
0.115000000000000	0.402350637004665\\ 
0.116000000000000	0.405464217425469\\ 
0.117000000000000	0.408257818364730\\ 
0.118000000000000	0.411372688487331\\ 
0.119000000000000	0.413905654042606\\ 
0.120000000000000	0.416745895015484\\ 
0.121000000000000	0.419475549469724\\ 
0.122000000000000	0.422013401226645\\ 
0.123000000000000	0.424470145260916\\ 
0.124000000000000	0.427411086443362\\ 
0.125000000000000	0.429874553738901\\ 
0.126000000000000	0.432385779534819\\ 
0.127000000000000	0.435000416922816\\ 
0.128000000000000	0.437543437448478\\ 
0.129000000000000	0.440196313677658\\ 
0.130000000000000	0.443281428602522\\ 
0.131000000000000	0.445825545922652\\ 
0.132000000000000	0.448205660752852\\ 
0.133000000000000	0.450697998378967\\ 
0.134000000000000	0.452934984578417\\ 
0.135000000000000	0.455152588916909\\ 
0.136000000000000	0.457875267799524\\ 
0.137000000000000	0.459968452264571\\ 
0.138000000000000	0.462013648067249\\ 
0.139000000000000	0.464152110663557\\ 
0.140000000000000	0.466637914448866\\ 
0.141000000000000	0.468779018577383\\ 
0.142000000000000	0.471171007055579\\ 
0.143000000000000	0.473581236737953\\ 
0.144000000000000	0.475642827814957\\ 
0.145000000000000	0.477675802298439\\ 
0.146000000000000	0.479427830245862\\ 
0.147000000000000	0.481577705092721\\ 
0.148000000000000	0.483479739459365\\ 
0.149000000000000	0.485321762500723\\ 
0.150000000000000	0.487253134029197\\ 
0.151000000000000	0.489402772596932\\ 
0.152000000000000	0.491389553226622\\ 
0.153000000000000	0.493254318521210\\ 
0.154000000000000	0.494797325731140\\ 
0.155000000000000	0.496583179617376\\ 
0.156000000000000	0.498380908426659\\ 
0.157000000000000	0.500044026587387\\ 
0.158000000000000	0.502165413692772\\ 
0.159000000000000	0.503805440287970\\ 
0.160000000000000	0.505543364639440\\ 
0.161000000000000	0.507272928936456\\ 
0.162000000000000	0.509284998937418\\ 
0.163000000000000	0.510912805872165\\ 
0.164000000000000	0.512857723656647\\ 
0.165000000000000	0.514569260880956\\ 
0.166000000000000	0.516525664752851\\ 
0.167000000000000	0.518330142559040\\ 
0.168000000000000	0.520040014096400\\ 
0.169000000000000	0.522105225484210\\ 
0.170000000000000	0.523912909384984\\ 
0.171000000000000	0.525584529568752\\ 
0.172000000000000	0.527111662023466\\ 
0.173000000000000	0.528923771921084\\ 
0.174000000000000	0.530670925958247\\ 
0.175000000000000	0.532028409139946\\ 
0.176000000000000	0.533762204138852\\ 
0.177000000000000	0.535295050349295\\ 
0.178000000000000	0.537167601956619\\ 
0.179000000000000	0.539037917951594\\ 
0.180000000000000	0.540765127504460\\ 
0.181000000000000	0.542595364092847\\ 
0.182000000000000	0.544026595541885\\ 
0.183000000000000	0.545360269000817\\ 
0.184000000000000	0.547022107407011\\ 
0.185000000000000	0.548220607683593\\ 
0.186000000000000	0.549388408028729\\ 
0.187000000000000	0.550835124266950\\ 
0.188000000000000	0.552730390064245\\ 
0.189000000000000	0.554357366135467\\ 
0.190000000000000	0.555967711589979\\ 
0.191000000000000	0.557570074226444\\ 
0.192000000000000	0.559209925252128\\ 
0.193000000000000	0.560817957274775\\ 
0.194000000000000	0.562222940087011\\ 
0.195000000000000	0.563675689950162\\ 
0.196000000000000	0.564907793195281\\ 
0.197000000000000	0.566322147402908\\ 
0.198000000000000	0.567621540862944\\ 
0.199000000000000	0.568997327652255\\ 
0.200000000000000	0.570450514607975\\
};
\addlegendentry{FARM};

\addplot [color=orange,solid,line width=1.5pt]
  table[row sep=crcr]{%
0	0.000802147119758213\\ 
0.00100000000000000	0.000802147119758213\\ 
0.00200000000000000	0.000802147119758213\\ 
0.00300000000000000	0.000824590030603750\\ 
0.00400000000000000	0.000886941278922702\\ 
0.00500000000000000	0.00104938178904459\\ 
0.00600000000000000	0.00149124990617857\\ 
0.00700000000000000	0.00190238436612961\\ 
0.00800000000000000	0.00252763278893600\\ 
0.00900000000000000	0.00319526894797862\\ 
0.0100000000000000	0.00427635286989677\\ 
0.0110000000000000	0.00588225406298143\\ 
0.0120000000000000	0.00738391557328765\\ 
0.0130000000000000	0.00845748154185275\\ 
0.0140000000000000	0.00967357447816065\\ 
0.0150000000000000	0.0110072379632224\\ 
0.0160000000000000	0.0124900477553343\\ 
0.0170000000000000	0.0138090597795119\\ 
0.0180000000000000	0.0152755353206728\\ 
0.0190000000000000	0.0167680161228316\\ 
0.0200000000000000	0.0182563673145176\\ 
0.0210000000000000	0.0198616953446937\\ 
0.0220000000000000	0.0216438602582486\\ 
0.0230000000000000	0.0236774215597886\\ 
0.0240000000000000	0.0254676416479721\\ 
0.0250000000000000	0.0274348812618276\\ 
0.0260000000000000	0.0296402279805518\\ 
0.0270000000000000	0.0316458588605874\\ 
0.0280000000000000	0.0339717346657345\\ 
0.0290000000000000	0.0363096264328794\\ 
0.0300000000000000	0.0390276094388541\\ 
0.0310000000000000	0.0411444268130186\\ 
0.0320000000000000	0.0434100843267502\\ 
0.0330000000000000	0.0456530522704005\\ 
0.0340000000000000	0.0481252073755671\\ 
0.0350000000000000	0.0506012562396236\\ 
0.0360000000000000	0.0526953315129588\\ 
0.0370000000000000	0.0551201080003486\\ 
0.0380000000000000	0.0575167966478801\\ 
0.0390000000000000	0.0602498177805075\\ 
0.0400000000000000	0.0628212822771282\\ 
0.0410000000000000	0.0654435609835266\\ 
0.0420000000000000	0.0680773705823135\\ 
0.0430000000000000	0.0706764530369060\\ 
0.0440000000000000	0.0733707249571486\\ 
0.0450000000000000	0.0758855999547832\\ 
0.0460000000000000	0.0783297828094283\\ 
0.0470000000000000	0.0809072256671832\\ 
0.0480000000000000	0.0837835578298770\\ 
0.0490000000000000	0.0863906961548620\\ 
0.0500000000000000	0.0894305217594497\\ 
0.0510000000000000	0.0920882403364204\\ 
0.0520000000000000	0.0953321005826812\\ 
0.0530000000000000	0.0983045391940057\\ 
0.0540000000000000	0.101382303723401\\ 
0.0550000000000000	0.104703867519567\\ 
0.0560000000000000	0.107922181858142\\ 
0.0570000000000000	0.111109125918239\\ 
0.0580000000000000	0.114442867964786\\ 
0.0590000000000000	0.117564719945398\\ 
0.0600000000000000	0.120892284495625\\ 
0.0610000000000000	0.124462039015576\\ 
0.0620000000000000	0.127606689453865\\ 
0.0630000000000000	0.130824162832686\\ 
0.0640000000000000	0.134021827624678\\ 
0.0650000000000000	0.137500933566087\\ 
0.0660000000000000	0.140579817393414\\ 
0.0670000000000000	0.144243820926179\\ 
0.0680000000000000	0.147970623140344\\ 
0.0690000000000000	0.150961465821720\\ 
0.0700000000000000	0.154406591210241\\ 
0.0710000000000000	0.158001060084819\\ 
0.0720000000000000	0.161451971854223\\ 
0.0730000000000000	0.164895405423817\\ 
0.0740000000000000	0.168213754196420\\ 
0.0750000000000000	0.171758661774734\\ 
0.0760000000000000	0.175200919991832\\ 
0.0770000000000000	0.178947668156309\\ 
0.0780000000000000	0.182281315511723\\ 
0.0790000000000000	0.185821836051005\\ 
0.0800000000000000	0.189463954310304\\ 
0.0810000000000000	0.193105103268430\\ 
0.0820000000000000	0.196510949627525\\ 
0.0830000000000000	0.199933633433812\\ 
0.0840000000000000	0.203459798634916\\ 
0.0850000000000000	0.207300964486712\\ 
0.0860000000000000	0.211120166641900\\ 
0.0870000000000000	0.215088800814128\\ 
0.0880000000000000	0.218722952507389\\ 
0.0890000000000000	0.222431467083153\\ 
0.0900000000000000	0.225815798357078\\ 
0.0910000000000000	0.229051325928047\\ 
0.0920000000000000	0.232415232213767\\ 
0.0930000000000000	0.236204328879625\\ 
0.0940000000000000	0.239932455906558\\ 
0.0950000000000000	0.243561155300787\\ 
0.0960000000000000	0.246987741861352\\ 
0.0970000000000000	0.250796599319658\\ 
0.0980000000000000	0.254368218619481\\ 
0.0990000000000000	0.257897978466815\\ 
0.100000000000000	0.261832374889448\\ 
0.101000000000000	0.265339382085004\\ 
0.102000000000000	0.268652137243192\\ 
0.103000000000000	0.272497931267916\\ 
0.104000000000000	0.276106173101563\\ 
0.105000000000000	0.279961462508994\\ 
0.106000000000000	0.283665395977039\\ 
0.107000000000000	0.287020488274944\\ 
0.108000000000000	0.290534750393877\\ 
0.109000000000000	0.294755447933931\\ 
0.110000000000000	0.298394244914437\\ 
0.111000000000000	0.302238257852465\\ 
0.112000000000000	0.306001120303995\\ 
0.113000000000000	0.310123371732017\\ 
0.114000000000000	0.314027599026205\\ 
0.115000000000000	0.317774294477875\\ 
0.116000000000000	0.321906972034728\\ 
0.117000000000000	0.325738466745293\\ 
0.118000000000000	0.329426390028327\\ 
0.119000000000000	0.333381762825937\\ 
0.120000000000000	0.337262424775397\\ 
0.121000000000000	0.341186952599484\\ 
0.122000000000000	0.344919522011831\\ 
0.123000000000000	0.348726458851454\\ 
0.124000000000000	0.352398033209733\\ 
0.125000000000000	0.356488143783997\\ 
0.126000000000000	0.360036629530118\\ 
0.127000000000000	0.363867322278964\\ 
0.128000000000000	0.367447543716243\\ 
0.129000000000000	0.370848931377109\\ 
0.130000000000000	0.374364597774399\\ 
0.131000000000000	0.377903975858896\\ 
0.132000000000000	0.381449702625709\\ 
0.133000000000000	0.385350722984064\\ 
0.134000000000000	0.388841461669617\\ 
0.135000000000000	0.392800140886315\\ 
0.136000000000000	0.396585948952460\\ 
0.137000000000000	0.400500775616646\\ 
0.138000000000000	0.404379621682419\\ 
0.139000000000000	0.408152168128012\\ 
0.140000000000000	0.412085416996458\\ 
0.141000000000000	0.416095774976395\\ 
0.142000000000000	0.420237540128020\\ 
0.143000000000000	0.424323483787082\\ 
0.144000000000000	0.428244788207058\\ 
0.145000000000000	0.432487425417380\\ 
0.146000000000000	0.436747410076304\\ 
0.147000000000000	0.440852626256316\\ 
0.148000000000000	0.445145304400231\\ 
0.149000000000000	0.449505461232361\\ 
0.150000000000000	0.454051679340013\\ 
0.151000000000000	0.458759127725686\\ 
0.152000000000000	0.462657655111307\\ 
0.153000000000000	0.466864330212135\\ 
0.154000000000000	0.471436318847632\\ 
0.155000000000000	0.475937016338162\\ 
0.156000000000000	0.480387328003500\\ 
0.157000000000000	0.484473964240788\\ 
0.158000000000000	0.488484932419775\\ 
0.159000000000000	0.492810843214418\\ 
0.160000000000000	0.496922837822256\\ 
0.161000000000000	0.501331346721289\\ 
0.162000000000000	0.505654419974528\\ 
0.163000000000000	0.510197042478033\\ 
0.164000000000000	0.514431794107564\\ 
0.165000000000000	0.518719677094200\\ 
0.166000000000000	0.522522872933786\\ 
0.167000000000000	0.526862613414720\\ 
0.168000000000000	0.531097330753233\\ 
0.169000000000000	0.535379840484346\\ 
0.170000000000000	0.539321283567052\\ 
0.171000000000000	0.543132964685153\\ 
0.172000000000000	0.546988264112868\\ 
0.173000000000000	0.551001765028006\\ 
0.174000000000000	0.555077701594277\\ 
0.175000000000000	0.559560480810963\\ 
0.176000000000000	0.563813621664790\\ 
0.177000000000000	0.567776260327303\\ 
0.178000000000000	0.572198609718860\\ 
0.179000000000000	0.576795633596796\\ 
0.180000000000000	0.581031362654660\\ 
0.181000000000000	0.585039143774362\\ 
0.182000000000000	0.589292376402914\\ 
0.183000000000000	0.593381605843264\\ 
0.184000000000000	0.597521230083965\\ 
0.185000000000000	0.601865520712384\\ 
0.186000000000000	0.606007733695470\\ 
0.187000000000000	0.609975095389242\\ 
0.188000000000000	0.614271053477710\\ 
0.189000000000000	0.618100037737352\\ 
0.190000000000000	0.622334525183457\\ 
0.191000000000000	0.625963970066671\\ 
0.192000000000000	0.629972876211124\\ 
0.193000000000000	0.633774153490358\\ 
0.194000000000000	0.637388194739260\\ 
0.195000000000000	0.641319483853409\\ 
0.196000000000000	0.645114510172927\\ 
0.197000000000000	0.648387814233524\\ 
0.198000000000000	0.652032177635044\\ 
0.199000000000000	0.655669230536859\\ 
0.200000000000000	0.659207731121657\\
};
\addlegendentry{3D-CODED};

\addplot [color=cool_purple,solid,line width=2.0pt]
  table[row sep=crcr]{%
0	0.0642524523601943\\ 
0.00100000000000000	0.0643169827091742\\ 
0.00200000000000000	0.0651902169056076\\ 
0.00300000000000000	0.0686161736585869\\ 
0.00400000000000000	0.0754401392276066\\ 
0.00500000000000000	0.0856373022904707\\ 
0.00600000000000000	0.103418318730889\\ 
0.00700000000000000	0.128432932607856\\ 
0.00800000000000000	0.157041742820527\\ 
0.00900000000000000	0.190914948641674\\ 
0.0100000000000000	0.231093049626370\\ 
0.0110000000000000	0.273448186145332\\ 
0.0120000000000000	0.310952134057575\\ 
0.0130000000000000	0.346931947252189\\ 
0.0140000000000000	0.383805704930008\\ 
0.0150000000000000	0.420685599632002\\ 
0.0160000000000000	0.453387352155528\\ 
0.0170000000000000	0.486293802150621\\ 
0.0180000000000000	0.517017802564783\\ 
0.0190000000000000	0.547092441772068\\ 
0.0200000000000000	0.574557572527802\\ 
0.0210000000000000	0.602555390459371\\ 
0.0220000000000000	0.628908081490797\\ 
0.0230000000000000	0.652906030057361\\ 
0.0240000000000000	0.675776971106321\\ 
0.0250000000000000	0.696601249417387\\ 
0.0260000000000000	0.715377272339607\\ 
0.0270000000000000	0.733113523217518\\ 
0.0280000000000000	0.747970502860068\\ 
0.0290000000000000	0.762665797616311\\ 
0.0300000000000000	0.775874552767092\\ 
0.0310000000000000	0.788331380165903\\ 
0.0320000000000000	0.800619183801736\\ 
0.0330000000000000	0.812196283589353\\ 
0.0340000000000000	0.823149244588301\\ 
0.0350000000000000	0.833172066991602\\ 
0.0360000000000000	0.842394366286626\\ 
0.0370000000000000	0.851233549737863\\ 
0.0380000000000000	0.859428981480892\\ 
0.0390000000000000	0.867266281068947\\ 
0.0400000000000000	0.875278688334529\\ 
0.0410000000000000	0.882720972744820\\ 
0.0420000000000000	0.889343537931969\\ 
0.0430000000000000	0.895751669585011\\ 
0.0440000000000000	0.901838677821200\\ 
0.0450000000000000	0.907449742569343\\ 
0.0460000000000000	0.913204030486048\\ 
0.0470000000000000	0.918309977235133\\ 
0.0480000000000000	0.923282790036484\\ 
0.0490000000000000	0.928141092768522\\ 
0.0500000000000000	0.932793249930471\\ 
0.0510000000000000	0.936880190117266\\ 
0.0520000000000000	0.940877852845818\\ 
0.0530000000000000	0.944862639604392\\ 
0.0540000000000000	0.948367358282521\\ 
0.0550000000000000	0.951581055811807\\ 
0.0560000000000000	0.954492627618061\\ 
0.0570000000000000	0.957525648228760\\ 
0.0580000000000000	0.960481990001090\\ 
0.0590000000000000	0.963327709629185\\ 
0.0600000000000000	0.965896472018374\\ 
0.0610000000000000	0.968562361693801\\ 
0.0620000000000000	0.971111197695477\\ 
0.0630000000000000	0.973633564821460\\ 
0.0640000000000000	0.975760411084178\\ 
0.0650000000000000	0.977785551391942\\ 
0.0660000000000000	0.979681283926156\\ 
0.0670000000000000	0.981434764729073\\ 
0.0680000000000000	0.982995489894577\\ 
0.0690000000000000	0.984385518631348\\ 
0.0700000000000000	0.985798102774474\\ 
0.0710000000000000	0.987012085078400\\ 
0.0720000000000000	0.988152428590568\\ 
0.0730000000000000	0.989199107275930\\ 
0.0740000000000000	0.990187647973416\\ 
0.0750000000000000	0.991013949418154\\ 
0.0760000000000000	0.991613683240774\\ 
0.0770000000000000	0.992265562293582\\ 
0.0780000000000000	0.992723288166834\\ 
0.0790000000000000	0.993204888326247\\ 
0.0800000000000000	0.993613128967720\\ 
0.0810000000000000	0.993906503208275\\ 
0.0820000000000000	0.994270352842243\\ 
0.0830000000000000	0.994568821958171\\ 
0.0840000000000000	0.994834188482299\\ 
0.0850000000000000	0.995054397120733\\ 
0.0860000000000000	0.995246250250991\\ 
0.0870000000000000	0.995424443435205\\ 
0.0880000000000000	0.995525090523650\\ 
0.0890000000000000	0.995677563837778\\ 
0.0900000000000000	0.995770269738343\\ 
0.0910000000000000	0.995928264511181\\ 
0.0920000000000000	0.996056357787095\\ 
0.0930000000000000	0.996099082631414\\ 
0.0940000000000000	0.996166606022010\\ 
0.0950000000000000	0.996178811563326\\ 
0.0960000000000000	0.996223861522788\\ 
0.0970000000000000	0.996258595443751\\ 
0.0980000000000000	0.996327875956554\\ 
0.0990000000000000	0.996436510426511\\ 
0.100000000000000	0.996515236707814\\ 
0.101000000000000	0.996607228300469\\ 
0.102000000000000	0.996617622230414\\ 
0.103000000000000	0.996670549878605\\ 
0.104000000000000	0.996770530790982\\ 
0.105000000000000	0.996805772106712\\ 
0.106000000000000	0.996840424844372\\ 
0.107000000000000	0.996885229523414\\ 
0.108000000000000	0.996907451333644\\ 
0.109000000000000	0.996917845263589\\ 
0.110000000000000	0.996942132477008\\ 
0.111000000000000	0.996975047986767\\ 
0.112000000000000	0.996995441583178\\ 
0.113000000000000	0.997019988508551\\ 
0.114000000000000	0.997055097546430\\ 
0.115000000000000	0.997065294344636\\ 
0.116000000000000	0.997101955825405\\ 
0.117000000000000	0.997114161366721\\ 
0.118000000000000	0.997136760837982\\ 
0.119000000000000	0.997161216777435\\ 
0.120000000000000	0.997161216777435\\ 
0.121000000000000	0.997171413575641\\ 
0.122000000000000	0.997171413575641\\ 
0.123000000000000	0.997183935488989\\ 
0.124000000000000	0.997217365193349\\ 
0.125000000000000	0.997217365193349\\ 
0.126000000000000	0.997229390205374\\ 
0.127000000000000	0.997229390205374\\ 
0.128000000000000	0.997229390205374\\ 
0.129000000000000	0.997241471877477\\ 
0.130000000000000	0.997253677418793\\ 
0.131000000000000	0.997266199332141\\ 
0.132000000000000	0.997289946886384\\ 
0.133000000000000	0.997301488899311\\ 
0.134000000000000	0.997336113382597\\ 
0.135000000000000	0.997336113382597\\ 
0.136000000000000	0.997348138394622\\ 
0.137000000000000	0.997360343935938\\ 
0.138000000000000	0.997372549477254\\ 
0.139000000000000	0.997384091490181\\ 
0.140000000000000	0.997384091490181\\ 
0.141000000000000	0.997384091490181\\ 
0.142000000000000	0.997396613403529\\ 
0.143000000000000	0.997419032345499\\ 
0.144000000000000	0.997431554258847\\ 
0.145000000000000	0.997443635930951\\ 
0.146000000000000	0.997455177943878\\ 
0.147000000000000	0.997455177943878\\ 
0.148000000000000	0.997478744968830\\ 
0.149000000000000	0.997478744968830\\ 
0.150000000000000	0.997478744968830\\ 
0.151000000000000	0.997491266882178\\ 
0.152000000000000	0.997491266882178\\ 
0.153000000000000	0.997491266882178\\ 
0.154000000000000	0.997491266882178\\ 
0.155000000000000	0.997491266882178\\ 
0.156000000000000	0.997503291894203\\ 
0.157000000000000	0.997503291894203\\ 
0.158000000000000	0.997503291894203\\ 
0.159000000000000	0.997503291894203\\ 
0.160000000000000	0.997514833907130\\ 
0.161000000000000	0.997526375920057\\ 
0.162000000000000	0.997537917932984\\ 
0.163000000000000	0.997574187400575\\ 
0.164000000000000	0.997586269072679\\ 
0.165000000000000	0.997586269072679\\ 
0.166000000000000	0.997586269072679\\ 
0.167000000000000	0.997598790986027\\ 
0.168000000000000	0.997611312899375\\ 
0.169000000000000	0.997611312899375\\ 
0.170000000000000	0.997611312899375\\ 
0.171000000000000	0.997611312899375\\ 
0.172000000000000	0.997611312899375\\ 
0.173000000000000	0.997611312899375\\ 
0.174000000000000	0.997611312899375\\ 
0.175000000000000	0.997611312899375\\ 
0.176000000000000	0.997611312899375\\ 
0.177000000000000	0.997634936584406\\ 
0.178000000000000	0.997634936584406\\ 
0.179000000000000	0.997647018256509\\ 
0.180000000000000	0.997647018256509\\ 
0.181000000000000	0.997647018256509\\ 
0.182000000000000	0.997647018256509\\ 
0.183000000000000	0.997647018256509\\ 
0.184000000000000	0.997647018256509\\ 
0.185000000000000	0.997659099928613\\ 
0.186000000000000	0.997659099928613\\ 
0.187000000000000	0.997671621841961\\ 
0.188000000000000	0.997671621841961\\ 
0.189000000000000	0.997671621841961\\ 
0.190000000000000	0.997671621841961\\ 
0.191000000000000	0.997671621841961\\ 
0.192000000000000	0.997671621841961\\ 
0.193000000000000	0.997671621841961\\ 
0.194000000000000	0.997683703514064\\ 
0.195000000000000	0.997683703514064\\ 
0.196000000000000	0.997696225427413\\ 
0.197000000000000	0.997708307099516\\ 
0.198000000000000	0.997708307099516\\ 
0.199000000000000	0.997720388771620\\ 
0.200000000000000	0.997720388771620\\
};
\addlegendentry{Ours};

\end{axis}
\end{tikzpicture}

%% file: sections/conclusions.tex
We proposed an alignment-based solution to the problem of shape completion from range scans. Different from most previous works, we focus on the setting where a complete shape is given, but is at a different pose than that of the scan. Our data-driven solution is based on learning the space of distortions, linking scans at various poses to whole shapes in other poses. As a result, at test time we can align unseen pairs of parts and whole shapes at different poses. In this paper we focused on human shapes. This is mainly due to the availability of rich data sources. That said, our method is not restricted to a particular class of shapes and in the future we plan to explore other shape classes. 
From Ancient Greek holistic philosophy, through modern psychology explanations of the human brain perception of shapes, we have demonstrated that computational matching procedures could benefit from the same axiomatic assumption stating that indeed {\it the whole is larger than the sum of its parts}. 

%% file: sections/Supplementary.tex
\noindent
In this appendix we provide
\begin{enumerate}
     \item Analysis of our network; We provide an ablation experiment introspecting the influence of the full shape $Q$ on the network reconstructions. Additionally, we provide robustness analysis of our trained network in Section \ref{method_analysis}.  
    \item Additional visualizations of the network reconstructions in Section \ref{additional_vis}.
    \item Visualizations of the dense correspondence results from the partial shape to the full shape in Section \ref{dense_corr}.
\end{enumerate}

\section{Analysis} \label{method_analysis}
\subsection{Comparison with a fixed template baseline}
\label{fixed_template}
As described in the main manuscript, in order to predict the completion of a partial shape $P$, our method requires a full reference shape $Q$ of the same subject in an arbitrary pose.
We motivate this setting by a requirement for a completion that is faithful to the subject shape. 
This is different from previous completion methods which can only approximate or hallucinate missing details.

Here we would like to support this claim experimentally, by comparing with a baseline which uses a fixed template. 
Specifically, instead of providing a full shape $Q$ of the same subject as the partial shape $P$, we provide a \textit{fixed} full template $T$ for all inputs. 

With this modification, the ablation network is trained with the triplets $\{(P_n, T, R_n)\}_{n=1}^N$, where $N$ is the size of the training set. At inference time, we use the same template $T$ to make a prediction for a given input part $P$. We chose the template to be the first subject from the FAUST Projections dataset, in its null pose. Both the original and the fixed-template networks were trained on the FAUST Projections training set, with identical parameters and for the same number of epochs, as described in Section $3.6$ in the paper. Table \ref{ablation} summarizes the prediction errors of both methods, Figure  \ref{fig:ablation_corr_curve} compares the partial correspondence results and Figure \ref{fig:ablation} shows visual comparison. The results clearly show the benefit of utilizing the shared geometry between the part and a full non-rigid observation of it. In particular, we receive a noticeable improvement in correspondence prediction as well as a lower reconstruction error across all metrics. Perhaps more importantly, Figure \ref{fig:ablation} demonstrates the main motivation of our framework: a completion that respects the fine details of the underlying shape. To further emphasize this effect, we magnify the face regions of each shape, showing the loss in detail achieved with the alternative training method.  

\begin{figure}[htbp]
    \centering
    \includegraphics[width = 0.22\linewidth]{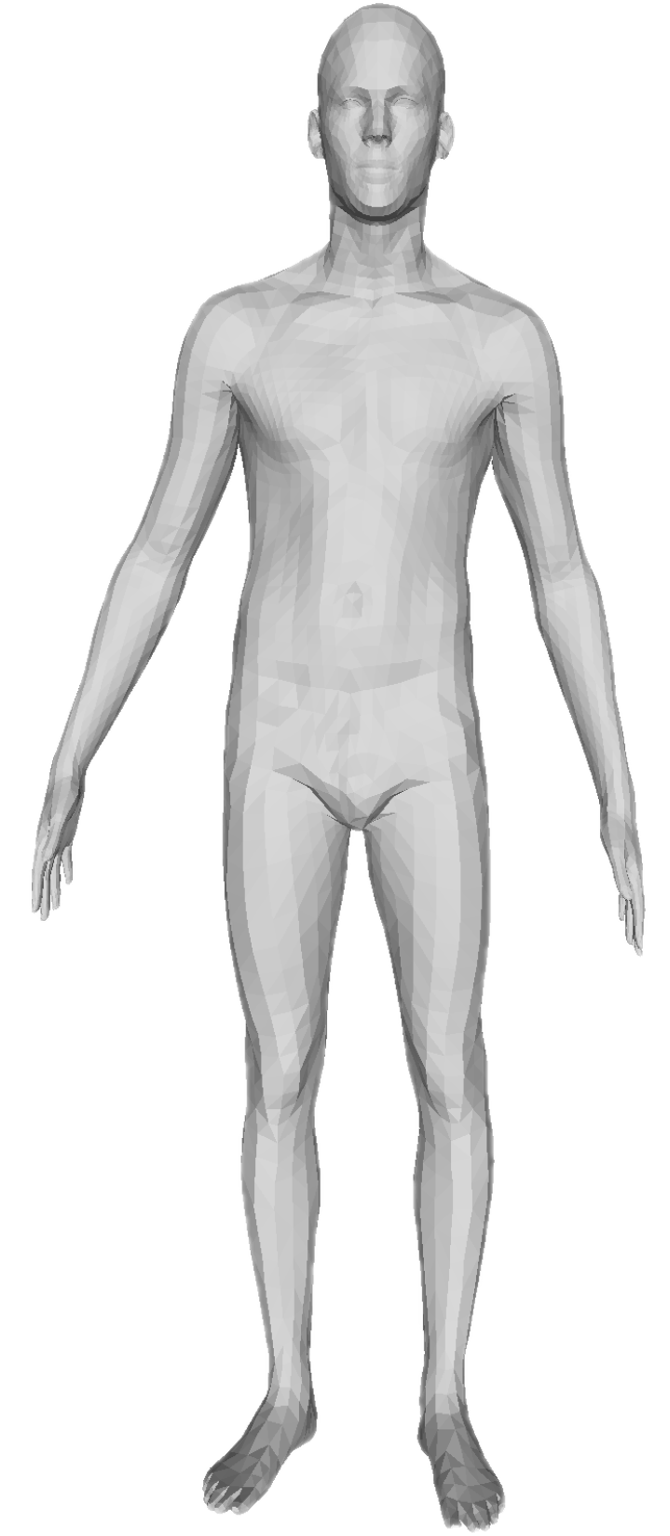}
    \caption{
    {Constant template used in ablation fixed-template experiment}}
    \label{fig:template}
\end{figure}
\FloatBarrier

Figure \ref{fig:ablation} implies how powerful our method is when it comes to the reconstruction of fine details, such as the facial structure and delicate body features. We verify that acquiring access to a full observation in inference time can significantly improve the reliability of the reconstruction for a network trained to utilize such information.
In the absence of this full observation at inference time, the ablation network can only utilize the input part and the acquired statistics of the training examples, encoded in the network weights. While this later information can be used for coarse completion, we evidence it is not sufficient for accurate completion. 

\begin{figure*}[htbp]
    \centering
    \includegraphics[width = 0.98\textwidth]{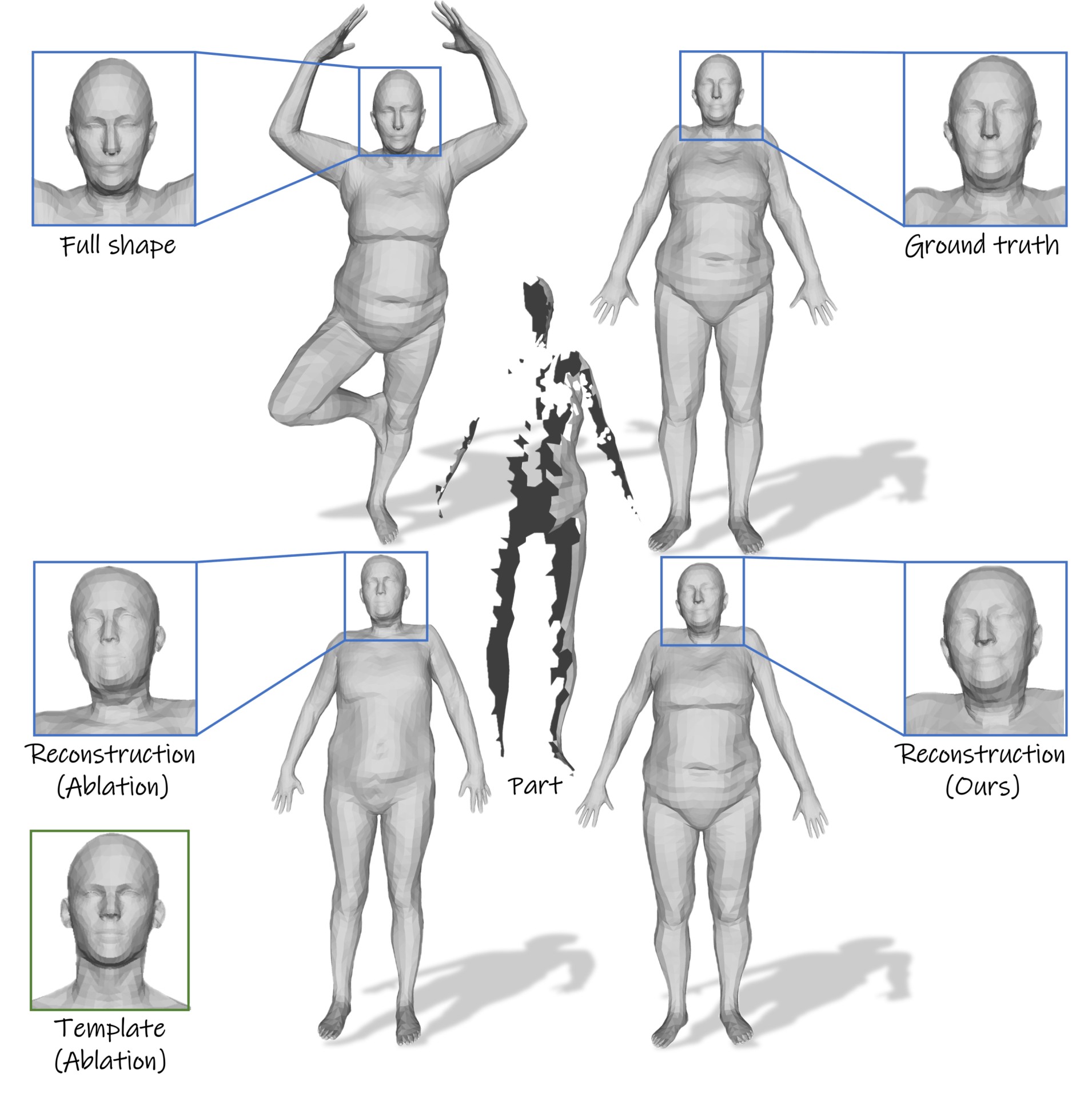}
    \caption{
    {Comparison with fixed-template ablation experiment.}}
    \label{fig:ablation}
\end{figure*}
\FloatBarrier

\input{tables/table_ablation_training_with_constant_template.tex}

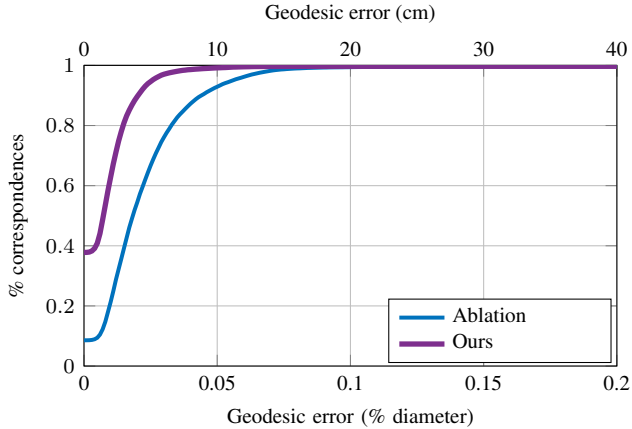
\begin{figure}[htbp]
	\centering
	\setlength\figureheight{4cm}
	\setlength\figurewidth{\linewidth}
	\input{./figures/Faust_Ablation_Geoerr.tikz}
\caption{
{Comparison with fixed-template ablation experiment.} Partial correspondence error evaluated on FAUST Projections dataset.}
\label{fig:ablation_corr_curve}
\end{figure}

\subsection{Robustness Analysis} \label{robustness_analysis}

We turn to analyze the robustness and stability of our proposed method, in hopes of shedding light of its possible applicability in real world conditions. Three specific aspects of the method were inspected empirically, each allowing for a realization of some non-optimal condition commonly found in real scans. The following experiments utilize a network trained over the FAUST train set. The realization is provided over a test-set of 200 single-view projected scans produced from 10 azimutal viewpoints around 2 human subjects exhibiting 10 different poses. The relevant full shapes were taken from the FAUST dataset, and are completely disjoint from our train set. Each scan $P$ is matched with all possible poses $Q$ of the same subject, achieving a total of 2000 inputs. We utilize a descriptive partial set of the evaluation metrics proposed in section 4.3 of the paper to evaluate each experiment. 

\paragraph{Residual Noise}

\begin{figure}[htbp]
    \centering
    \includegraphics[width = \linewidth]{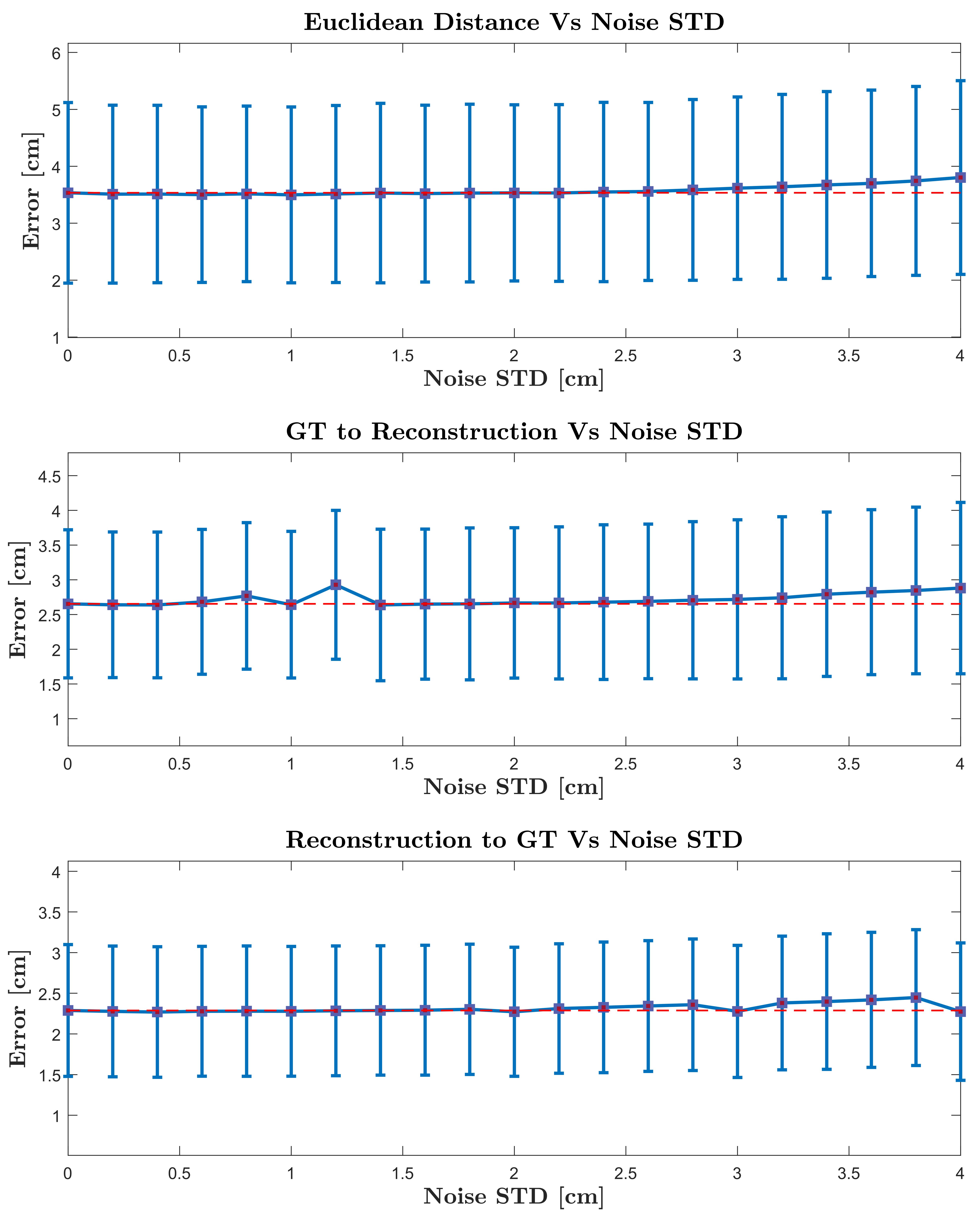}
    \caption{\textbf{Robustness to Noise.} Three reconstruction metrics evaluated on completions originating from corrupted partial shapes with varying levels of additive white Gaussian noise. A baseline with the evaluation realized with no noise is marked with a dashed red line.}
    \label{fig:noise_analysis}
\end{figure}

In this experiment, we attempt to emulate various artifacts commonly found in segmented depth scans. We corrupt the vertices of each partial input shape with various degrees of additive white Gaussian noise, with standard deviations in the range [0-4] cm. The corrupted partial shapes are fed to the network, together with the full shapes. Averaged reconstruction statistics are displayed graphically in Figure \ref{fig:noise_analysis}. As apparent from the figure, the method accuracy only slightly declines with the increase of the noise. 

\paragraph{Downsampling}

\begin{figure}[htbp]
    \centering
    \includegraphics[width = 0.87\linewidth]{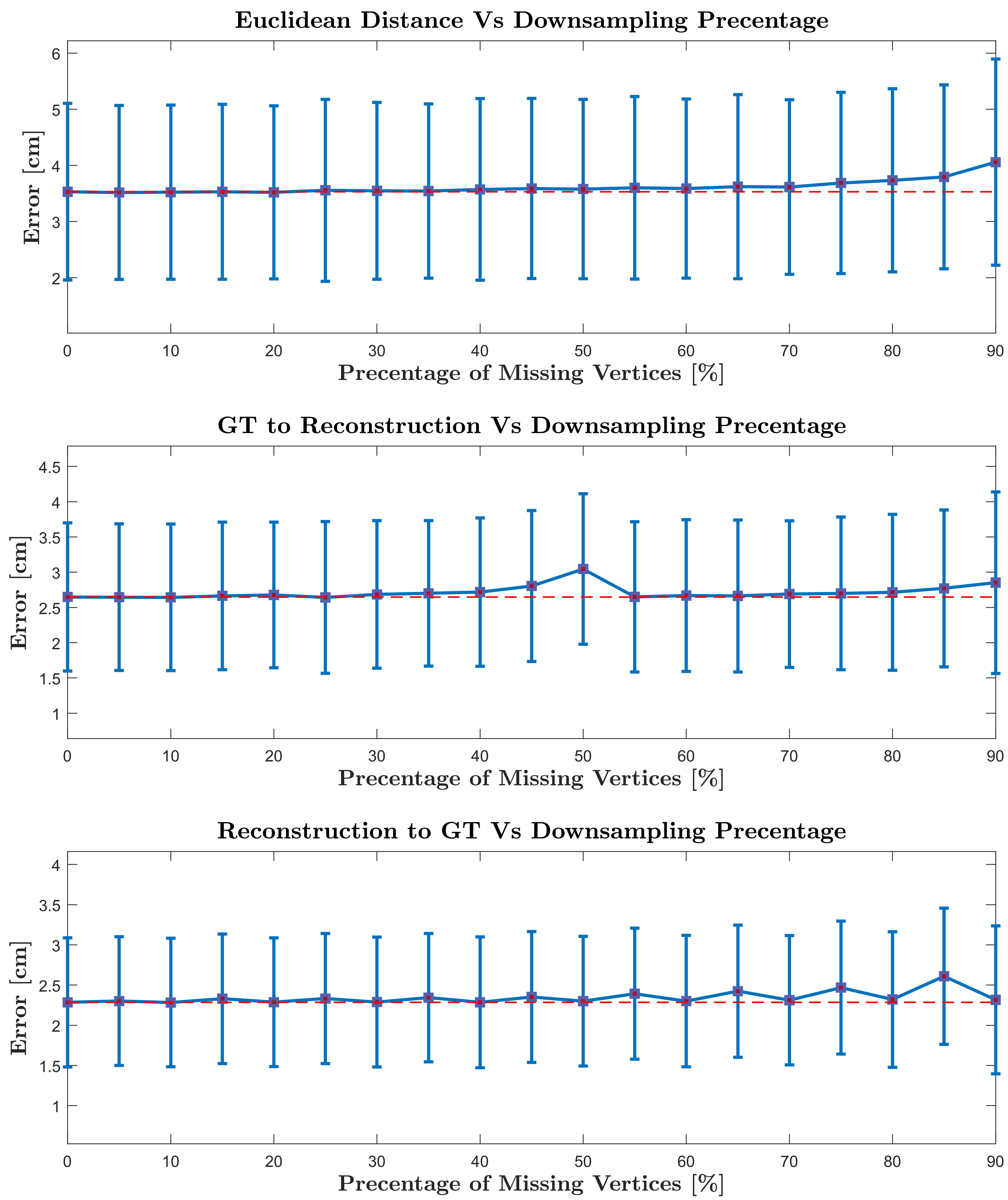}
    \caption{\textbf{Robustness to Downsampling.}  Three reconstruction metrics evaluated on completions originating from decimated partial shapes with varying levels of vertex erasure.  A baseline with the evaluation realized with no decimation is marked with a dashed red line.}
    \label{fig:downsamp}
\end{figure}

We address the network's ability to infer on partial shapes with decreasing degrees of resolution. For each partial shape in the mentioned test set, we decimate at random some percentage of the existing vertices, and infer on the resultant set. As can be seen in Figure \ref{fig:downsamp}, even under a majority decimation of the vertices, the proposed network is able to recover well the ground truth shape. 

\paragraph{Projection Angle}

\begin{figure}[htbp]
    \centering
    \includegraphics[width = 0.87\linewidth]{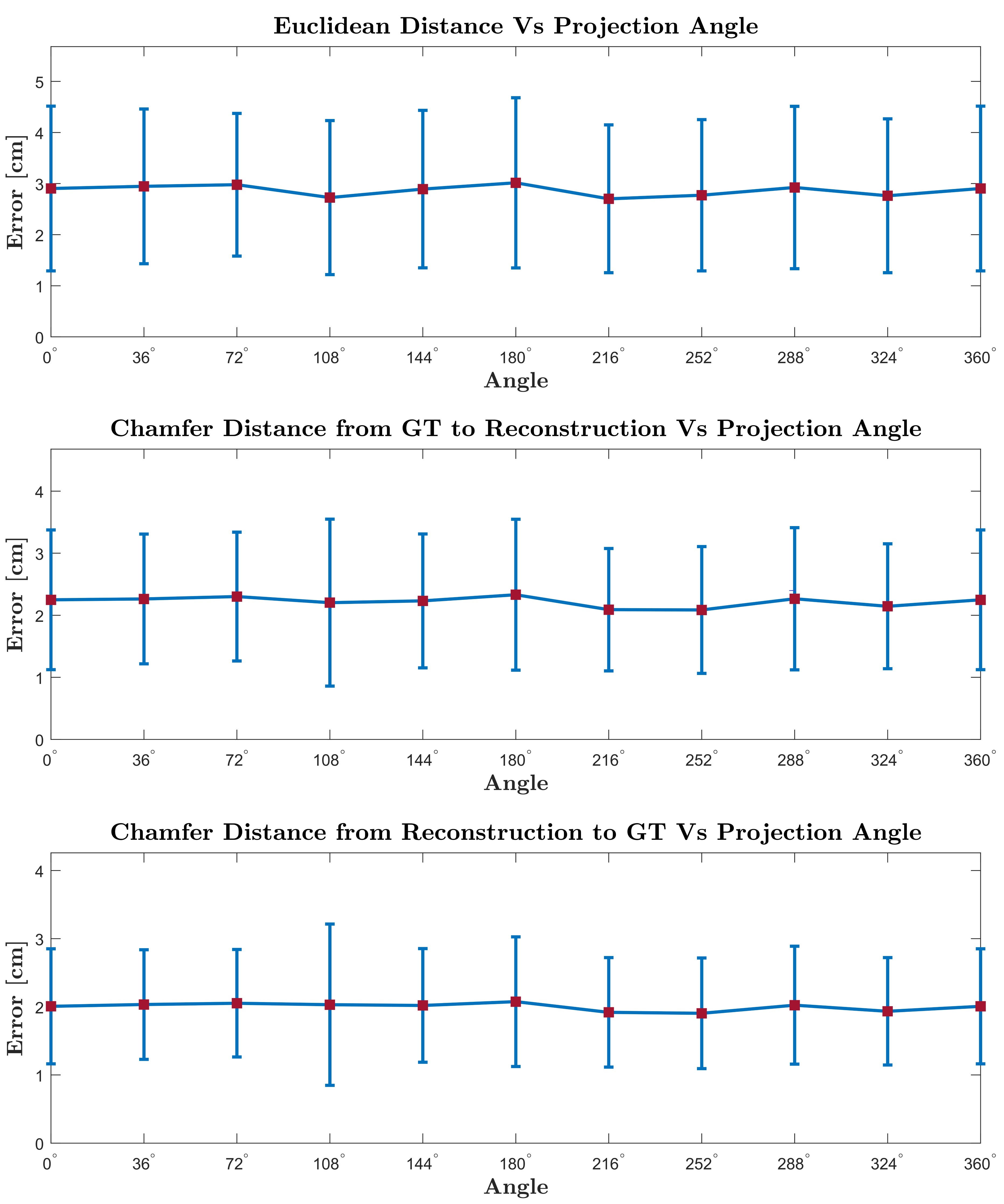}
    \caption{\textbf{Robustness to Projection Angle.} Three reconstruction metrics evaluated on different groups of the test-set, partitioned by the projection angle. We note a close to uniform distribution over the different angles, attributing to a azimutal invariancy.}
    \label{fig:angle_inv}
\end{figure}

Finally, we examine the dependency of our network to the projection angle. We note that due to the different projection angles and poses, it is not unreasonable that some angles hold a higher degree of information relevant for reconstruction than others. Ideally, we would like to enable the network a reliable reconstruction at every angle, regardless if the information seen is the back, front or sides of a shape. We partition the 2000 completions received over the test set into their corresponding projection angles, and accumulate the errors over each partition. The result is displayed in Figure \ref{fig:angle_inv}. The received error distribution is close to uniform, attributing to the method's azimutal invariancy.

\section{Additional Visualizations} \label{additional_vis}
\begin{figure*}[!htbp]
    \centering
    \includegraphics[width =\textwidth]{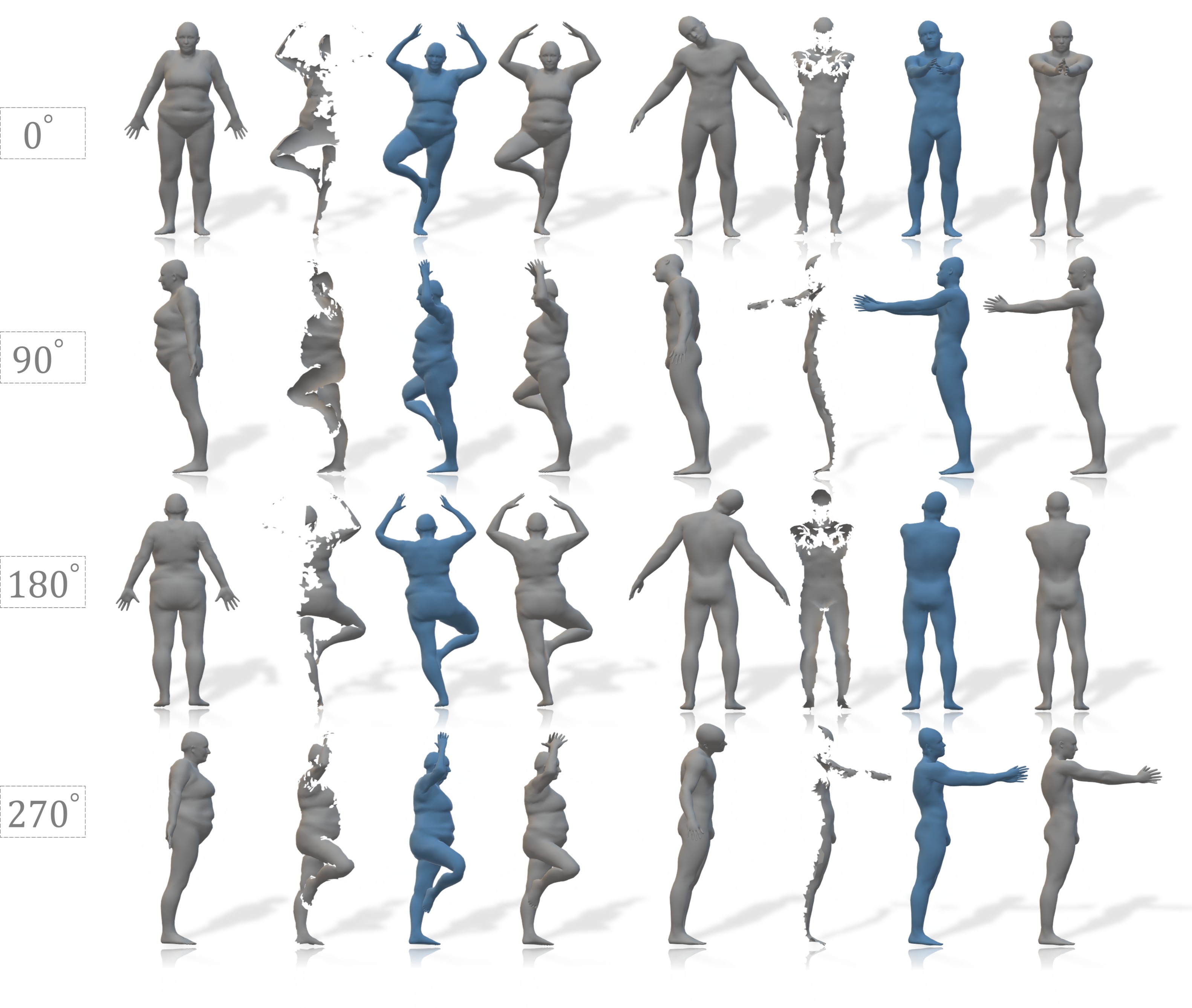}
    \caption{\textbf{Predicted completions, FAUST Projections.} Each column shows a completion for a different subject, while each row provides a different perspective on the reconstructed 3D model. From left to right: full input shape $Q$, input part $P$, predicted completion $F_{\theta(P,Q)}(Q)$, ground truth completion $R$.}
    \label{fig:faust_reconstructions_1}
\end{figure*}

\begin{figure*}[!htbp]
    \centering
    \includegraphics[width =\textwidth]{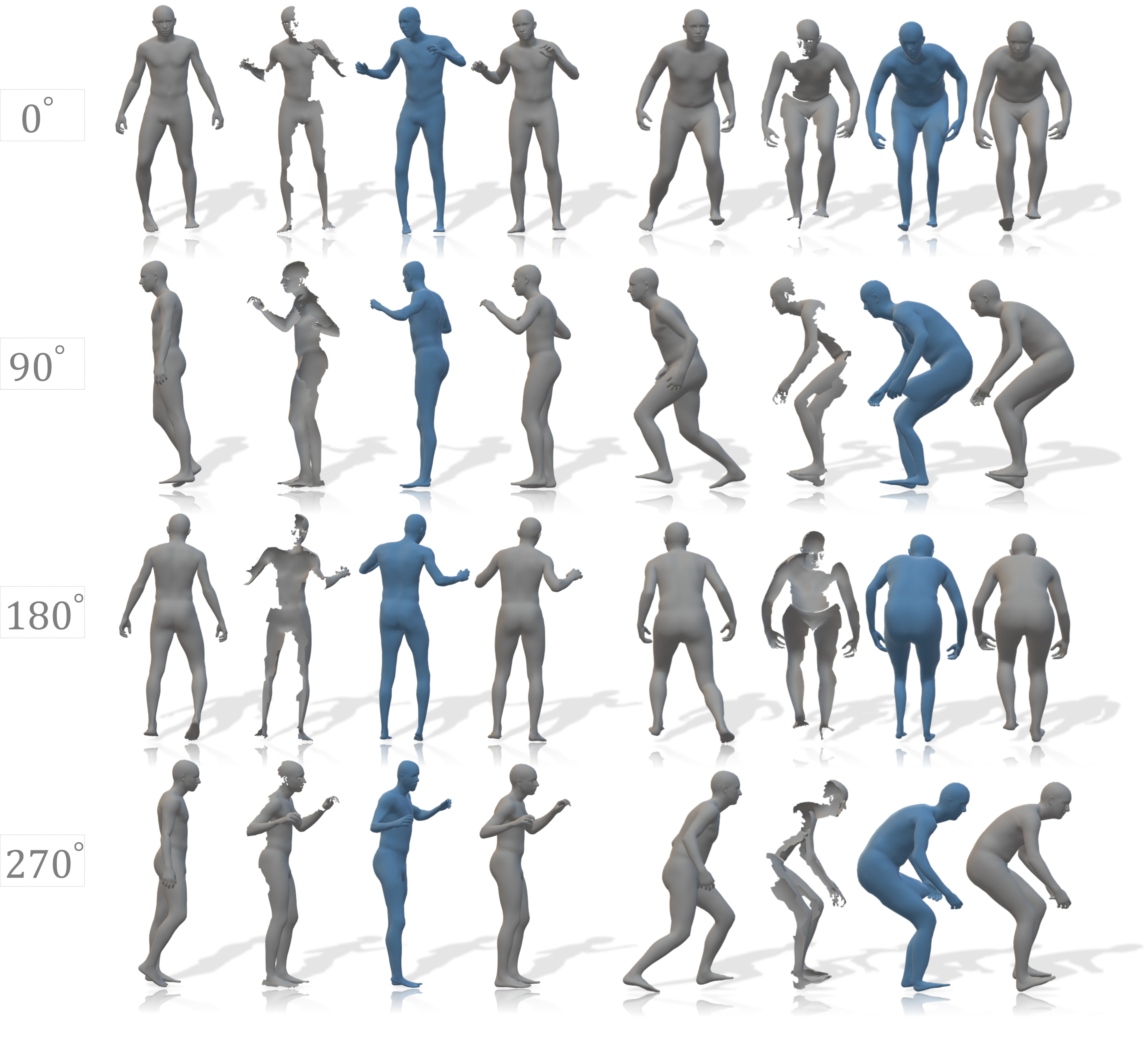}
    \caption{\textbf{Predicted completions, AMASS Projections.} Each column shows a completion for a different subject, while each row provides a different perspective on the reconstructed 3D model. From left to right: full input shape $Q$, input part $P$, predicted completion $F_{\theta(P,Q)}(Q)$, ground truth completion $R$.}
    \label{fig:amass}
\end{figure*}

Here we provide additional reconstructions that were not included in the main paper in order to save space. 
Figure \ref{fig:faust_reconstructions_1} and Figure \ref{fig:amass} visualize our network predictions for examples from \textbf{FAUST Projections} and \textbf{AMASS Projections}, respectively.

\section{Non-Rigid partial correspondence} \label{dense_corr}
\begin{figure*}[!htbp]
    \centering
    \includegraphics[width =\textwidth]{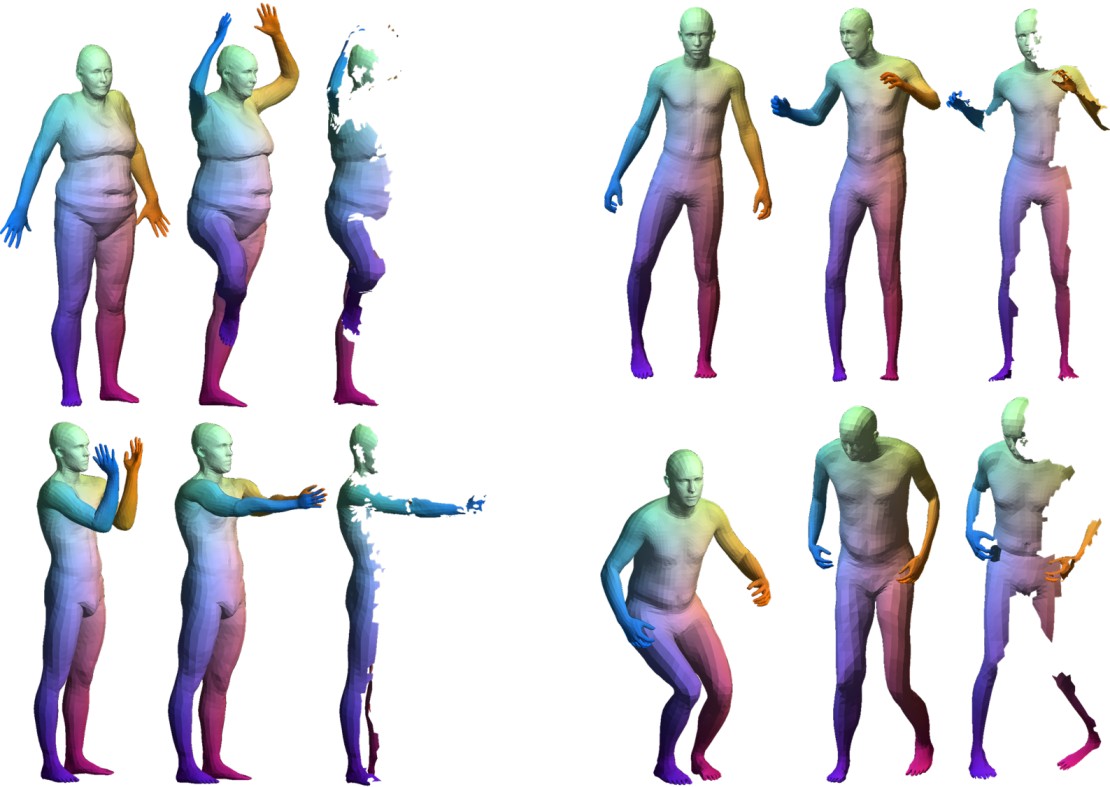}
    \caption{\textbf{Non-Rigid partial correspondence.} Left and right columns show the dense correspondence for FAUST Projections and AMASS Projections, respectively. From left to right: full input shape $Q$, our network completion $F_{\theta(P,Q)}(Q)$ and partial input shape $P$. Corresponding points are indicated by the same color.}
    \label{fig:correspondence}
\end{figure*}

    Figure \ref{fig:correspondence} visualizes the dense correspondence between the input partial and full shape. As explained in the paper, we achieve this by using the network reconstruction as a proxy; For every point in the partial shape we calculate the nearest neighbor point in the reconstruction allowing us a recovery of a mapping between the partial shape to the reconstructed shape, which is by construction also the mapping between the part and the full input shape. In Section 4.5 of the paper we evaluated the predicted correspondence numerically for FAUST Projections and AMASS Projections datasets, providing geodesic error graphs for both, in Figure 4 and Figure 5, respectively. For completion, we show the results also qualitatively here.

%% file: tables/table_ablation_training_with_constant_template.tex
\begin{table*}[!htbp]
\centering
\small

\begin{tabular}{ l@{\hskip 0.01\textwidth}c@{\hskip 0.01\textwidth}c@{\hskip 0.01\textwidth}c@{\hskip 0.01\textwidth}c@{\hskip 0.01\textwidth}c@{\hskip 0.01\textwidth}c@{\hskip 0.01\textwidth}c  }
    \hline\hline
    Error & Euclidean distance& Volumetric err. & Directional Chamfer distance & Directional Chamfer distance & Full Chamfer\\     
          & GT and reconstruction [cm] & mean $\pm$ std [\%] &   GT to reconstruction [cm]& reconstruction to GT [cm] & distance [cm]\\ \hline    
          
    Ablation &   $3.74$        &   $17.63 \pm 7.41$     &   $3.00$    &   $2.32$        &   $5.32$\\

    {\bf Ours}  & $\textbf{2.94}$ 	& $\textbf{7.05} \pm \textbf{3.45}$           &  $\textbf{2.42}$ &  $\textbf{1.95}$   & $\textbf{4.37}$ \\ \hline \hline
  \end{tabular}
  \vspace{2mm}

\caption{\small \textbf{Comparison with Fixed-Template Ablation Experiment.} We evaluate our method against an ablation experiment, repeating exactly the same training except of one significant difference: instead of providing the full shape $Q_n$ as described in the main paper, we provided a \textit{constant} full template $T$ in each of the training examples $\{(P_n, T, R_n)\}_{n=1}^N$. The template $T$ is used in inference as well, to predict the completion of a given input part $P$. We report the prediction errors on FAUST test set, while both networks were trained on FAUST train set. The first and second rows summarize the ablation errors and our method errors, respectively.}
\label{ablation}
\end{table*}

%% file: figures/Faust_Ablation_Geoerr.tikz
%
%
\definecolor{cool_yellow}{rgb}{0.92900,0.69400,0.12500}%
\definecolor{cool_green}{rgb}{0.46600,0.67400,0.18800}%
\definecolor{cool_purple}{rgb}{0.49400,0.18400,0.55600}%
\definecolor{cool_blue}{rgb}{0.00000,0.44700,0.74100}%
\definecolor{cool_red}{rgb}{0.85000,0.32500,0.09800}%
\begin{tikzpicture}

\pgfplotsset{compat=newest} 

\tikzstyle{every node}=[font=\footnotesize]

\begin{axis}[%
width=0.85\figurewidth,
height=\figureheight,
scale only axis,
xmin=0,
xmax=0.2,
xlabel={Geodesic error (cm)},
xtick = {0,0.05,0.1,0.15,0.2},
xticklabels = {0,10,20,30,40},
axis x line*=top,
axis y line=none
]
\addplot [color=white,opacity=0.0,line width=1pt,forget plot]
  table[row sep=crcr]{%
0	0.0661103047895501\\
0.203030303030303	0.998846153846154\\
};
\end{axis}

\begin{axis}[%
width=0.85\figurewidth,
height=\figureheight,
scale only axis,
xmin=0,
xmax=0.2,
xlabel style={align=center,text width=5cm},
xlabel={Geodesic error (\% diameter)},
xtick={0,0.05,0.1,0.15,0.2},
xticklabels = {0,0.05,0.1,0.15,0.2},
xmajorgrids,
ymin=0,
ymax=1,
ytick={0,0.2,0.4,0.6,0.8,1},
ylabel={\% correspondences},
ymajorgrids,
axis background/.style={fill=white},
legend style={
	at={(0.99,0.01)},
	anchor=south east,
	legend cell align=left,
	align=left,
	text width=5.50em,
	text height=1ex
}
]

\addplot [color=cool_blue,solid,line width=1.5pt]
  table[row sep=crcr]{%
0	0.0856879634739778 \\ 
0.00100000000000000	0.0856879634739778\\ 
0.00200000000000000	0.0859295108020625\\ 
0.00300000000000000	0.0868806458278340\\ 
0.00400000000000000	0.0891833447999408\\ 
0.00500000000000000	0.0941669329271949\\ 
0.00600000000000000	0.104576790137347\\ 
0.00700000000000000	0.122286097570066\\ 
0.00800000000000000	0.147297722530039\\ 
0.00900000000000000	0.180000856154461\\ 
0.0100000000000000	0.212906081298411\\ 
0.0110000000000000	0.249072430059029\\ 
0.0120000000000000	0.287764491561124\\ 
0.0130000000000000	0.322325463023545\\ 
0.0140000000000000	0.356699289677300\\ 
0.0150000000000000	0.391655417994085\\ 
0.0160000000000000	0.426283575885930\\ 
0.0170000000000000	0.460699741983603\\ 
0.0180000000000000	0.490349696048099\\ 
0.0190000000000000	0.519176075399275\\ 
0.0200000000000000	0.544839310010761\\ 
0.0210000000000000	0.572587829942176\\ 
0.0220000000000000	0.596223406422589\\ 
0.0230000000000000	0.620429095539572\\ 
0.0240000000000000	0.644065717881771\\ 
0.0250000000000000	0.666806972763854\\ 
0.0260000000000000	0.687951371101290\\ 
0.0270000000000000	0.707999576223959\\ 
0.0280000000000000	0.726411175328683\\ 
0.0290000000000000	0.745051567839119\\ 
0.0300000000000000	0.760734722166617\\ 
0.0310000000000000	0.775737201319091\\ 
0.0320000000000000	0.789114659922664\\ 
0.0330000000000000	0.803241933561144\\ 
0.0340000000000000	0.815167185599951\\ 
0.0350000000000000	0.827148125067904\\ 
0.0360000000000000	0.837140617798174\\ 
0.0370000000000000	0.846100097887930\\ 
0.0380000000000000	0.855379473515312\\ 
0.0390000000000000	0.864215151849447\\ 
0.0400000000000000	0.872765316481725\\ 
0.0410000000000000	0.880363553001510\\ 
0.0420000000000000	0.887816011586986\\ 
0.0430000000000000	0.894369967985016\\ 
0.0440000000000000	0.899717487953568\\ 
0.0450000000000000	0.904760110359555\\ 
0.0460000000000000	0.910546735370847\\ 
0.0470000000000000	0.915390745084399\\ 
0.0480000000000000	0.920201389284909\\ 
0.0490000000000000	0.924943544128239\\ 
0.0500000000000000	0.929023212539067\\ 
0.0510000000000000	0.933190019853568\\ 
0.0520000000000000	0.937612833326736\\ 
0.0530000000000000	0.941060427728888\\ 
0.0540000000000000	0.944007411111578\\ 
0.0550000000000000	0.947323146783668\\ 
0.0560000000000000	0.950649661098167\\ 
0.0570000000000000	0.953946784900828\\ 
0.0580000000000000	0.956532974755202\\ 
0.0590000000000000	0.959094087761855\\ 
0.0600000000000000	0.961915504621017\\ 
0.0610000000000000	0.964646852063403\\ 
0.0620000000000000	0.967219239567396\\ 
0.0630000000000000	0.969741652655687\\ 
0.0640000000000000	0.971680792597772\\ 
0.0650000000000000	0.973691972358228\\ 
0.0660000000000000	0.975681519560107\\ 
0.0670000000000000	0.977306673535596\\ 
0.0680000000000000	0.978987328284732\\ 
0.0690000000000000	0.980527288719651\\ 
0.0700000000000000	0.982063131648418\\ 
0.0710000000000000	0.983468080462171\\ 
0.0720000000000000	0.984567506703428\\ 
0.0730000000000000	0.985483705683263\\ 
0.0740000000000000	0.986412264759013\\ 
0.0750000000000000	0.986951426777541\\ 
0.0760000000000000	0.987783557468405\\ 
0.0770000000000000	0.988372651487349\\ 
0.0780000000000000	0.988955146207560\\ 
0.0790000000000000	0.989506884833432\\ 
0.0800000000000000	0.989911785225583\\ 
0.0810000000000000	0.990175101326693\\ 
0.0820000000000000	0.990659487242171\\ 
0.0830000000000000	0.991061340635872\\ 
0.0840000000000000	0.991511311250606\\ 
0.0850000000000000	0.991686008651092\\ 
0.0860000000000000	0.991975770304439\\ 
0.0870000000000000	0.992197673610969\\ 
0.0880000000000000	0.992481140070621\\ 
0.0890000000000000	0.992692036046279\\ 
0.0900000000000000	0.992899741144854\\ 
0.0910000000000000	0.992991774484504\\ 
0.0920000000000000	0.993286530690496\\ 
0.0930000000000000	0.993373063553588\\ 
0.0940000000000000	0.993502674786648\\ 
0.0950000000000000	0.993559822251592\\ 
0.0960000000000000	0.993681326357399\\ 
0.0970000000000000	0.993710446916514\\ 
0.0980000000000000	0.993892896642219\\ 
0.0990000000000000	0.993951137760448\\ 
0.100000000000000	0.994152584792315\\ 
0.101000000000000	0.994327596849841\\ 
0.102000000000000	0.994417686892659\\ 
0.103000000000000	0.994537887769814\\ 
0.104000000000000	0.994567008328929\\ 
0.105000000000000	0.994655980854657\\ 
0.106000000000000	0.994655980854657\\ 
0.107000000000000	0.994803194498614\\ 
0.108000000000000	0.994861435616844\\ 
0.109000000000000	0.995015567643484\\ 
0.110000000000000	0.995073808761714\\ 
0.111000000000000	0.995202522043439\\ 
0.112000000000000	0.995260763161668\\ 
0.113000000000000	0.995412481479875\\ 
0.114000000000000	0.995505958679852\\ 
0.115000000000000	0.995715918116288\\ 
0.116000000000000	0.995774159234517\\ 
0.117000000000000	0.995873751957128\\ 
0.118000000000000	0.995961113634472\\ 
0.119000000000000	0.996051141148124\\ 
0.120000000000000	0.996109382266354\\ 
0.121000000000000	0.996231980025446\\ 
0.122000000000000	0.996348462261905\\ 
0.123000000000000	0.996404164426548\\ 
0.124000000000000	0.996433284985663\\ 
0.125000000000000	0.996532877708273\\ 
0.126000000000000	0.996632470430883\\ 
0.127000000000000	0.996725947630860\\ 
0.128000000000000	0.996842429867319\\ 
0.129000000000000	0.996958912103778\\ 
0.130000000000000	0.996958912103778\\ 
0.131000000000000	0.997090291221811\\ 
0.132000000000000	0.997151198176349\\ 
0.133000000000000	0.997151198176349\\ 
0.134000000000000	0.997151198176349\\ 
0.135000000000000	0.997186434258097\\ 
0.136000000000000	0.997215554817211\\ 
0.137000000000000	0.997273795935441\\ 
0.138000000000000	0.997309032017188\\ 
0.139000000000000	0.997338152576303\\ 
0.140000000000000	0.997338152576303\\ 
0.141000000000000	0.997396393694533\\ 
0.142000000000000	0.997425514253647\\ 
0.143000000000000	0.997479560226384\\ 
0.144000000000000	0.997505293628340\\ 
0.145000000000000	0.997505293628340\\ 
0.146000000000000	0.997505293628340\\ 
0.147000000000000	0.997505293628340\\ 
0.148000000000000	0.997505293628340\\ 
0.149000000000000	0.997534414187455\\ 
0.150000000000000	0.997604886350950\\ 
0.151000000000000	0.997604886350950\\ 
0.152000000000000	0.997604886350950\\ 
0.153000000000000	0.997604886350950\\ 
0.154000000000000	0.997604886350950\\ 
0.155000000000000	0.997663127469180\\ 
0.156000000000000	0.997663127469180\\ 
0.157000000000000	0.997663127469180\\ 
0.158000000000000	0.997663127469180\\ 
0.159000000000000	0.997663127469180\\ 
0.160000000000000	0.997663127469180\\ 
0.161000000000000	0.997698363550927\\ 
0.162000000000000	0.997698363550927\\ 
0.163000000000000	0.997698363550927\\ 
0.164000000000000	0.997698363550927\\ 
0.165000000000000	0.997698363550927\\ 
0.166000000000000	0.997698363550927\\ 
0.167000000000000	0.997698363550927\\ 
0.168000000000000	0.997698363550927\\ 
0.169000000000000	0.997727484110042\\ 
0.170000000000000	0.997727484110042\\ 
0.171000000000000	0.997727484110042\\ 
0.172000000000000	0.997727484110042\\ 
0.173000000000000	0.997727484110042\\ 
0.174000000000000	0.997727484110042\\ 
0.175000000000000	0.997778950913954\\ 
0.176000000000000	0.997836470711332\\ 
0.177000000000000	0.997836470711332\\ 
0.178000000000000	0.997836470711332\\ 
0.179000000000000	0.997836470711332\\ 
0.180000000000000	0.997836470711332\\ 
0.181000000000000	0.997836470711332\\ 
0.182000000000000	0.997836470711332\\ 
0.183000000000000	0.997836470711332\\ 
0.184000000000000	0.997836470711332\\ 
0.185000000000000	0.997836470711332\\ 
0.186000000000000	0.997836470711332\\ 
0.187000000000000	0.997865591270447\\ 
0.188000000000000	0.997865591270447\\ 
0.189000000000000	0.997865591270447\\ 
0.190000000000000	0.997865591270447\\ 
0.191000000000000	0.997865591270447\\ 
0.192000000000000	0.997891324672403\\ 
0.193000000000000	0.997917058074358\\ 
0.194000000000000	0.997917058074358\\ 
0.195000000000000	0.997917058074358\\ 
0.196000000000000	0.997946178633473\\ 
0.197000000000000	0.997946178633473\\ 
0.198000000000000	0.997946178633473\\ 
0.199000000000000	0.997946178633473\\ 
0.200000000000000	0.997975299192588\\
};
\addlegendentry{Ablation};

\addplot [color=cool_purple,solid,line width=2.0pt]
  table[row sep=crcr]{%
0.0	0.377951611998809\\
0.00100000000000000	0.378036033373903\\
0.00200000000000000	0.379167035155197\\
0.00300000000000000	0.382959244917291\\
0.00400000000000000	0.391535548561057\\
0.00500000000000000	0.408818636076324\\
0.00600000000000000	0.440863484994918\\
0.00700000000000000	0.486986791528920\\
0.00800000000000000	0.535073908425617\\
0.00900000000000000	0.583559115754113\\
0.0100000000000000	0.628341576956260\\
0.0110000000000000	0.672363368054937\\
0.0120000000000000	0.711595990977030\\
0.0130000000000000	0.748568452371760\\
0.0140000000000000	0.780159395085799\\
0.0150000000000000	0.808631424390558\\
0.0160000000000000	0.832118187383844\\
0.0170000000000000	0.851277095584864\\
0.0180000000000000	0.868261232595781\\
0.0190000000000000	0.883536241669241\\
0.0200000000000000	0.896601465316566\\
0.0210000000000000	0.908714600320218\\
0.0220000000000000	0.920553477737559\\
0.0230000000000000	0.931021438637458\\
0.0240000000000000	0.939638114344266\\
0.0250000000000000	0.946225505167257\\
0.0260000000000000	0.952889616305034\\
0.0270000000000000	0.958273704100909\\
0.0280000000000000	0.962960560438036\\
0.0290000000000000	0.967412618263656\\
0.0300000000000000	0.970573097325085\\
0.0310000000000000	0.972973217130731\\
0.0320000000000000	0.975261318147548\\
0.0330000000000000	0.977066744598635\\
0.0340000000000000	0.978870588760588\\
0.0350000000000000	0.980507384313727\\
0.0360000000000000	0.982060616490203\\
0.0370000000000000	0.983439454599471\\
0.0380000000000000	0.984334182555086\\
0.0390000000000000	0.985531718779241\\
0.0400000000000000	0.986060555942572\\
0.0410000000000000	0.986892094228688\\
0.0420000000000000	0.987611217343257\\
0.0430000000000000	0.988165374986298\\
0.0440000000000000	0.988850890393063\\
0.0450000000000000	0.989426205301674\\
0.0460000000000000	0.989884274719578\\
0.0470000000000000	0.990394589394571\\
0.0480000000000000	0.990940553344350\\
0.0490000000000000	0.991447205640969\\
0.0500000000000000	0.991881074568725\\
0.0510000000000000	0.992129842323015\\
0.0520000000000000	0.992563320439331\\
0.0530000000000000	0.992820604078365\\
0.0540000000000000	0.993358694946519\\
0.0550000000000000	0.993881761627720\\
0.0560000000000000	0.994384668270388\\
0.0570000000000000	0.994534066140127\\
0.0580000000000000	0.994825304733737\\
0.0590000000000000	0.995047304739619\\
0.0600000000000000	0.995316663843333\\
0.0610000000000000	0.995475507034166\\
0.0620000000000000	0.995648943162214\\
0.0630000000000000	0.995906392914526\\
0.0640000000000000	0.996073859127964\\
0.0650000000000000	0.996126268820448\\
0.0660000000000000	0.996238812226639\\
0.0670000000000000	0.996238812226639\\
0.0680000000000000	0.996332840307932\\
0.0690000000000000	0.996426517555212\\
0.0700000000000000	0.996487318217477\\
0.0710000000000000	0.996517080122239\\
0.0720000000000000	0.996569784122132\\
0.0730000000000000	0.996597271752699\\
0.0740000000000000	0.996665149206377\\
0.0750000000000000	0.996744528780451\\
0.0760000000000000	0.996771082152730\\
0.0770000000000000	0.996771082152730\\
0.0780000000000000	0.996771082152730\\
0.0790000000000000	0.996798569783296\\
0.0800000000000000	0.996798569783296\\
0.0810000000000000	0.996798569783296\\
0.0820000000000000	0.996824303185252\\
0.0830000000000000	0.996824303185252\\
0.0840000000000000	0.996824303185252\\
0.0850000000000000	0.996824303185252\\
0.0860000000000000	0.996824303185252\\
0.0870000000000000	0.996824303185252\\
0.0880000000000000	0.996824303185252\\
0.0890000000000000	0.996824303185252\\
0.0900000000000000	0.996824303185252\\
0.0910000000000000	0.996850856557530\\
0.0920000000000000	0.996850856557530\\
0.0930000000000000	0.996850856557530\\
0.0940000000000000	0.996850856557530\\
0.0950000000000000	0.996876589959485\\
0.0960000000000000	0.996876589959485\\
0.0970000000000000	0.996876589959485\\
0.0980000000000000	0.996876589959485\\
0.0990000000000000	0.996902323361441\\
0.100000000000000	0.996902323361441\\
0.101000000000000	0.996902323361441\\
0.102000000000000	0.996902323361441\\
0.103000000000000	0.996902323361441\\
0.104000000000000	0.996902323361441\\
0.105000000000000	0.996902323361441\\
0.106000000000000	0.996902323361441\\
0.107000000000000	0.996902323361441\\
0.108000000000000	0.996902323361441\\
0.109000000000000	0.996902323361441\\
0.110000000000000	0.996902323361441\\
0.111000000000000	0.996902323361441\\
0.112000000000000	0.996902323361441\\
0.113000000000000	0.996902323361441\\
0.114000000000000	0.996902323361441\\
0.115000000000000	0.996902323361441\\
0.116000000000000	0.996902323361441\\
0.117000000000000	0.996902323361441\\
0.118000000000000	0.996902323361441\\
0.119000000000000	0.996902323361441\\
0.120000000000000	0.996902323361441\\
0.121000000000000	0.996902323361441\\
0.122000000000000	0.996902323361441\\
0.123000000000000	0.996902323361441\\
0.124000000000000	0.996902323361441\\
0.125000000000000	0.996902323361441\\
0.126000000000000	0.996902323361441\\
0.127000000000000	0.996902323361441\\
0.128000000000000	0.996902323361441\\
0.129000000000000	0.996902323361441\\
0.130000000000000	0.996902323361441\\
0.131000000000000	0.996902323361441\\
0.132000000000000	0.996902323361441\\
0.133000000000000	0.996902323361441\\
0.134000000000000	0.996902323361441\\
0.135000000000000	0.996902323361441\\
0.136000000000000	0.996902323361441\\
0.137000000000000	0.996902323361441\\
0.138000000000000	0.996902323361441\\
0.139000000000000	0.996902323361441\\
0.140000000000000	0.996902323361441\\
0.141000000000000	0.996902323361441\\
0.142000000000000	0.996902323361441\\
0.143000000000000	0.996902323361441\\
0.144000000000000	0.996902323361441\\
0.145000000000000	0.996902323361441\\
0.146000000000000	0.996902323361441\\
0.147000000000000	0.996902323361441\\
0.148000000000000	0.996902323361441\\
0.149000000000000	0.996902323361441\\
0.150000000000000	0.996902323361441\\
0.151000000000000	0.996902323361441\\
0.152000000000000	0.996902323361441\\
0.153000000000000	0.996902323361441\\
0.154000000000000	0.996902323361441\\
0.155000000000000	0.996902323361441\\
0.156000000000000	0.996902323361441\\
0.157000000000000	0.996902323361441\\
0.158000000000000	0.996902323361441\\
0.159000000000000	0.996902323361441\\
0.160000000000000	0.996902323361441\\
0.161000000000000	0.996902323361441\\
0.162000000000000	0.996902323361441\\
0.163000000000000	0.996902323361441\\
0.164000000000000	0.996902323361441\\
0.165000000000000	0.996902323361441\\
0.166000000000000	0.996902323361441\\
0.167000000000000	0.996902323361441\\
0.168000000000000	0.996902323361441\\
0.169000000000000	0.996902323361441\\
0.170000000000000	0.996902323361441\\
0.171000000000000	0.996902323361441\\
0.172000000000000	0.996902323361441\\
0.173000000000000	0.996902323361441\\
0.174000000000000	0.996902323361441\\
0.175000000000000	0.996902323361441\\
0.176000000000000	0.996902323361441\\
0.177000000000000	0.996902323361441\\
0.178000000000000	0.996902323361441\\
0.179000000000000	0.996902323361441\\
0.180000000000000	0.996902323361441\\
0.181000000000000	0.996902323361441\\
0.182000000000000	0.996902323361441\\
0.183000000000000	0.996902323361441\\
0.184000000000000	0.996902323361441\\
0.185000000000000	0.996902323361441\\
0.186000000000000	0.996902323361441\\
0.187000000000000	0.996902323361441\\
0.188000000000000	0.996902323361441\\
0.189000000000000	0.996902323361441\\
0.190000000000000	0.996902323361441\\
0.191000000000000	0.996902323361441\\
0.192000000000000	0.996902323361441\\
0.193000000000000	0.996902323361441\\
0.194000000000000	0.996902323361441\\
0.195000000000000	0.996902323361441\\
0.196000000000000	0.996902323361441\\
0.197000000000000	0.996902323361441\\
0.198000000000000	0.996929810992008\\
0.199000000000000	0.996929810992008\\
0.200000000000000	0.996929810992008\\
};
\addlegendentry{Ours};

\end{axis}
\end{tikzpicture}